\documentclass{article}

\usepackage[preprint]{neurips_2025}

\usepackage[utf8]{inputenc} 
\usepackage[T1]{fontenc}    
\usepackage{hyperref}       
\usepackage{url}            
\usepackage{booktabs}       
\usepackage{amsfonts}       
\usepackage{nicefrac}       
\usepackage{microtype}      
\usepackage{xcolor}         

\newcommand{\model}{GraphDPO}

\newcommand{\task}{UKGE}
\usepackage{amsmath}
\usepackage{amssymb}
\usepackage{bm}

\usepackage{multirow}
\usepackage{graphicx}
\usepackage{float}
\usepackage{subfigure}
\usepackage{color}
\usepackage{wrapfig}
\usepackage{amsmath}
\newtheorem{theorem}{Theorem}

\usepackage[ruled,vlined]{algorithm2e}
\usepackage[absolute,overlay]{textpos}
\usepackage{authblk}

\title{Unlearning of Knowledge Graph Embedding via Preference Optimization}

\author[1]{Jiajun Liu}
\author[1,2*]{Wenjun Ke}
\author[1,2*]{Peng Wang}
\author[3]{Yao He}
\author[1]{Ziyu Shang}
\author[1]{Guozheng Li}
\author[1]{Zijie Xu}
\author[4]{Ke Ji}
\affil[1]{School of Computer Science and Engineering, Southeast University}
\affil[2]{Key Laboratory of New Generation Artificial Intelligence Technology and Its Interdisciplinary Applications (Southeast University), Ministry of Education}
\affil[3]{School of Institute of Collaborate Innovation, University of Macau}
\affil[4]{The Chinese University of Hongkong, Shenzhen}
\affil[ ]{\{jiajliu, kewenjun, pwang, ziyus1999, gzli, zijiexu\}@seu.edu.cn, mc46477@um.edu.mo, keji@link.cuhk.edu.cn}

\begin{document}

\maketitle

\begin{abstract}
Existing knowledge graphs (KGs) inevitably contain outdated or erroneous knowledge that needs to be removed from knowledge graph embedding (KGE) models. 
To address this challenge, knowledge unlearning can be applied to eliminate specific information while preserving the integrity of the remaining knowledge in KGs. 
Existing unlearning methods can generally be categorized into exact unlearning and approximate unlearning. 
However, exact unlearning requires high training costs while approximate unlearning faces two issues when applied to KGs due to the inherent connectivity of triples: 
(1) It fails to fully remove targeted information, as forgetting triples can still be inferred from remaining ones. 
(2) It focuses on local data for specific removal, which weakens the remaining knowledge in the forgetting boundary. 
To address these issues, we propose \model, a novel approximate unlearning framework based on direct preference optimization (DPO). 
Firstly, to effectively remove forgetting triples, we reframe unlearning as a preference optimization problem, where the model is trained by DPO to prefer reconstructed alternatives over the original forgetting triples. 
This formulation penalizes reliance on forgettable knowledge, mitigating incomplete forgetting caused by KG connectivity. 
Moreover, we introduce an out-boundary sampling strategy to construct preference pairs with minimal semantic overlap, weakening the connection between forgetting and retained knowledge. 
Secondly, to preserve boundary knowledge, we introduce a boundary recall mechanism that replays and distills relevant information both within and across time steps. 
We construct eight unlearning datasets across four popular KGs with varying unlearning rates. 
Experiments show that \model\ outperforms state-of-the-art baselines by up to 10.1\% in $MRR_{Avg}$ and 14.0\% in $MRR_{F1}$. 
Further analysis confirms that \model\ more effectively removes target knowledge while preserving surrounding context. 
\end{abstract}

\vspace{-4mm}
\section{Introduction}

\renewcommand{\thefootnote}{}
\footnotetext{*Corresponding author.}

Knowledge Graph Embedding (KGE)~\cite{wang2017knowledge,rossi2021knowledge} aims to embed entities and relations in knowledge graphs (KGs)~\cite{dong2014knowledge} into low-dimensional vectors, which is a crucial method for many knowledge-driven applications, such as question answering~\cite{bordes2014open}, semantic search~\cite{berant-liang-2014-semantic}, and information retrieval for large language models (LLMs)~\cite{barmettler2025conceptformer}. 
However, training data in KGs may contain incorrect and outdated information that needs to be removed~\cite{said2023survey}. 
For instance, nearly 5\% of the total knowledge in YAGO~\cite{10.1145/1242572.1242667} KG is incorrect~\cite{HOFFART201328}. 
Besides, vast outdated knowledge becomes irrelevant or incorrect due to changes in the real world~\cite{cheng2024editing}. 
For example, Wikidata KG has been edited more than 2 billion times from 2012 to 2024~\footnote{https://www.wikidata.org/w/index.php?title=Wikidata:Statistics}, which means up to 2 billion outdated pieces of knowledge should be forgotten.

\begin{wrapfigure}{htr}{0.5\textwidth}
  \centering
  \vspace{-4mm}
  \includegraphics[width=0.48\textwidth]{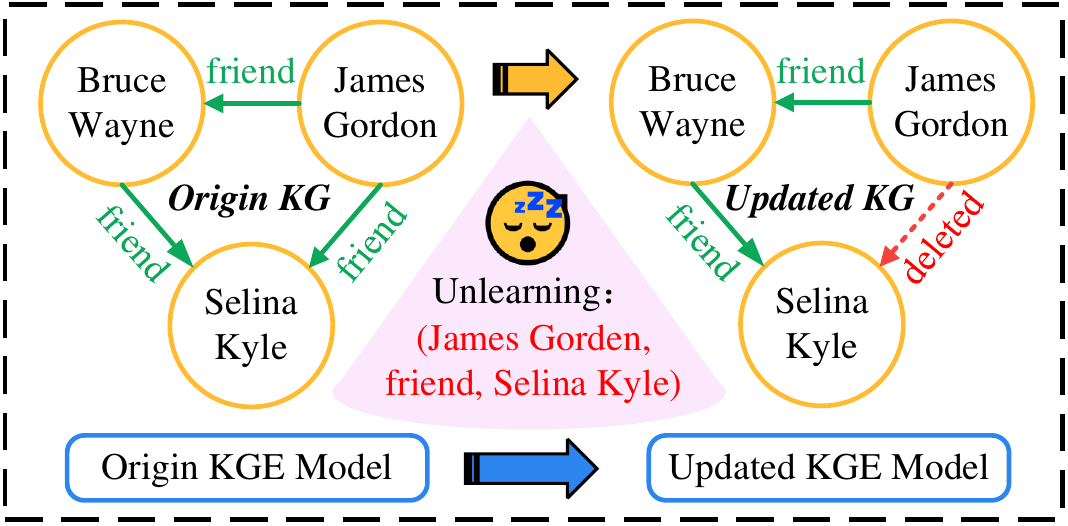}
  \vspace{-2mm}
  \caption{Illustration of unlearning in knowledge graph embedding (\task). The forgetting knowledge is (\textit{James Gordon}, \textit{friend}, \textit{Selina Kyle}), while other knowledge retains in the updated KG.}
  \vspace{-6mm}
  \label{intro}
\end{wrapfigure}

One promising approach to address these challenges is knowledge unlearning, which involves deleting specific knowledge from models~\cite{bourtoule2021machine}. 
However, existing knowledge unlearning methods are primarily designed for LLMs~\cite{eldan2023s} and computer vision models (CVMs)~\cite{wang2024machine}, with limited attention paid to unlearning in KGE models. 
Furthermore, existing KGE methods that leverage LLMs~\cite{cheng2024editing} mainly focus on the addition and modification of knowledge, overlooking the knowledge unlearning in KGs. 
Additionally, these LLM-based approaches tend to incur significant computational costs and require large-scale retraining, limiting their scalability and efficiency. 
In this paper, we study the \underline{\textbf{U}}nlearning of \underline{\textbf{K}}nowledge \underline{\textbf{G}}raph \underline{\textbf{E}}mbedding (\textbf{\task}) task, which aims to forget incorrect or outdated knowledge in pre-trained KGE models. 
Compared with methods based on LLMs, \task\  provides a more lightweight alternative without LLMs in scenarios where computational resources are limited. 
The crucial factor for the success of \task\ lies in effectively removing unlearned knowledge while preserving the remaining knowledge. 
As illustrated in Figure~\ref{intro}, while the triple \textit{(James Gordon, friend, Selina Kyle)} needs to be deleted, other triples such as \textit{(James Gordon, friend, Bruce Wayne)} should remain. 
Existing unlearning methods can be generally categorized into exact unlearning~\cite{golatkar2020eternal} and approximate unlearning~\cite{thudi2022unrolling}. 
Exact unlearning methods retrain the entire dataset to forget the specific knowledge~\cite{yan2022arcane}, thus coming with huge training costs.

To mitigate this drawback, approximate unlearning~\cite{wu2022puma} updates parameters using only forgetting data, thereby reducing training expenses. 
While approximate unlearning has made significant progress, it still suffers from the following issues when directly applied to \task\ due to the interconnected nature of KGs. 
Firstly, it fails to fully remove forgetting knowledge, as forgetting triples can still be inferred from the remaining ones. 
Secondly, it focuses on local data for specific removal, which weakens remaining knowledge in the forgetting boundary.

To address the above issues, we reframe unlearning as a preference optimization problem and propose a novel unlearning framework for knowledge \textbf{Graph} embedding with \textbf{DPO}, named \textbf{\model}, which efficiently forgets unwanted knowledge and preserves remaining knowledge. 
Specifically, the optimization objective for negative samples through preference optimization aligns with the objective of unlearning. 
Additionally, it allows for efficient sampling of positive samples on the graph structure by leveraging the graph connectivity features. 
Therefore, the preference optimization method is well-suited for addressing the challenge of forgetting learning on graph structures. 
Firstly, to mitigate the incomplete forgetting caused by the structural connectivity in KGs, preference optimization is modeled in such a way that forgetting triples are assigned larger penalty scores, thus allowing forgetting knowledge to be efficiently forgotten. 
Moreover, we design an out-boundary sampling strategy to construct alternative triples that are structurally distant from the forgetting triples, thus reducing their semantic and relational overlap with the retained knowledge. 
Then the DPO algorithm trains the model to prefer reconstructed alternatives over the original forgetting triples, suppressing their influence and preventing indirect reinforcement through retained structures. 
Secondly, to tackle the integrity issue of remaining knowledge in forgetting boundary, \model\ introduces a boundary recall mechanism to ensure minimal destruction of remaining knowledge. 
In detail, we replay and distill remaining knowledge at the forgetting boundary to preserve it within a single time step and across multiple time steps, respectively.

To validate the effectiveness of our method, we reorganize four commonly used datasets, FB15K-237~\cite{dettmers2018convolutional},  WN18RR~\cite{toutanova2015representing}, CoDEx-L~\cite{safavi2020codex}, and Yago3-10~\cite{mahdisoltani2013yago3}, and construct eight unlearning datasets with 10\% and 20\% unlearning rates, specifically for the novel UKGE task. 
Experimental results show that \model\ achieves optimal results in $MRR_{Avg}$ and $MRR_{F1}$ on most datasets compared to other strong approximate baselines. 
Our main contributions can be divided as follows: 
\begin{itemize}
  \item We propose \model, a novel \task\ framework that reformulates unlearning as preference optimization. By combining DPO with a graph-aware out-boundary sampling strategy, \model\ effectively reduces inference leakage from remaining knowledge.

  \item We design a boundary recall mechanism that explicitly preserves knowledge near the forgetting boundary. It combines knowledge replay and distillation to retain relevant context during the forgetting process, ensuring minimal degradation of remaining knowledge. 

  \item We construct and release eight benchmark datasets for \task\ across four standard KGs with varying unlearning rates. Extensive experiments show that \model\ consistently outperforms strong approximate unlearning baselines, achieving gains of up to 10.1\% in $MRR_{Avg}$ and 14.0\% in $MRR_{F1}$, while maintaining high training efficiency. 
\end{itemize}

\vspace{-2mm}
\section{Related Work}
\paragraph{Knowledge Unlearning.} Knowledge unlearning aims to remove incorrect and outdated knowledge from machine learning models~\cite{wang2024machine}. 
Existing unlearning methods can be categorized into exact and approximate approaches. 
Exact unlearning methods~\cite{golatkar2020eternal,nguyen2022survey} require re-training the model using all reserved data, resulting in significant training costs. 
To mitigate this issue, approximate unlearning methods~\cite{thudi2022unrolling,wu2022puma} update models using only the forgetting datasets, with little or no use of the reserved datasets, thereby reducing training time~\cite{foster2024fast,cha2024learning}. 
Recent works utilize schema~\cite{xiao2025knowledge} or meta-learning~\cite{xu2024learn} to forget knowledge in KGE models, however, they struggle to effectively forget while preserving remaining knowledge~\cite{bordes2013translating,trouillon2016complex,sun2019rotate} due to the inherent connectivity of KGs. 
\paragraph{Preserence Optimization.} Preference optimization seeks to align LLMs with human preferences and values~\cite{bai2022training}, and can be categorized into online and offline methods~\cite{meng2024simpo}. 
Online algorithms incorporate reinforcement learning with supervised fine-tuning and policy optimization, which are inherently challenging to optimize~\cite{schulman2017proximal}. 
To solve these issues, offline algorithms like DPO~\cite{rafailov2024direct} directly compare different decision sequences to optimize models, resulting in more efficient performance~\cite{azar2024general}. 
As alignment techniques leverage both positive and negative samples, offline algorithms are well-suited for selectively forgetting and remaining knowledge. 
In this paper, we demonstrate the effectiveness of applying DPO for unlearning in KGs. 
More detailed related work list in Appendix~\ref{appendix:related-work}.

\vspace{-2mm}
\section{Methodology}
\subsection{Preliminary and Problem Statement}
\paragraph{Knowledge Graph.}
A knowledge graph (KG) is denoted as $\mathcal{G} = \{\mathcal{E}, \mathcal{R}, \mathcal{T}\}$, where $\mathcal{E}$, $\mathcal{R}$, and $\mathcal{T}$ denote the set of entities, relations, and triples, respectively. 
A triple can be denoted as $(h, r, t)$, where $h \in \mathcal{E}$, $r \in \mathcal{R}$, and $t \in \mathcal{E}$ denote the head entity, the relation and the tail entity, respectively.

\paragraph{Knowledge Graph Embedding.} 
Knowledge graph embedding (KGE) embeds the entities and relations in KGs into low-dimensional vector space $\mathbb{R}^{d}$, where $d$ defines the embedding dimension. 
Specifically, KGE embeds the $h$, $r$, and $t$ in each triple to $\mathbf{h} \in \mathbb{R}^{d}$, $\mathbf{r} \in \mathbb{R}^{d}$, and $\mathbf{t} \in \mathbb{R}^{d}$, respectively. 
A KGE model can be defined as $\mathcal{M} = \{\mathbf{E}, \mathbf{R}, f\}$, where $\mathbf{E}$, $\mathbf{R}$ denote the set of embeddings of entities and relations, respectively, and $f$ denotes the score function for triples. 
The optimization objective for KGE is to minimize $\mathbb{E}_{(h, r, t) \sim \mathcal{T}, (h', r', t') \sim \overline{\mathcal{T}}} [f(h', r', t') - f(h, r, t)]$, where $\overline{\mathcal{T}}$ denotes the set of negative triples.

\paragraph{Unlearning in Knowledge Graph Embedding.}
The set of the training dataset for a KG $\mathcal{G}$ can be defined as $\mathcal{D}_{t} = \{ (h^{(i)}, r^{(i)}, t^{(i)}) \}_{i=1}^{N_{T}}$, where $N_{T}$ denotes the number of training triples and $\mathcal{D}_{t} \subseteq \mathcal{T}$. 
The set of the forgetting dataset can be defined as $\mathcal{D}_{f} \subset \mathcal{D}_{t}$, and the set of the remaining dataset can be defined as $\mathcal{D}_{r} = \mathcal{D}_{t} \backslash \mathcal{D}_{f}$. 
For a pre-trained KGE model $\mathcal{M}_{pre}$, unlearning in knowledge graph embedding (\task) aims to update $\mathbf{E}$ and $\mathbf{R}$ in $\mathcal{M}_{pre}$ to remove knowledge in $\mathcal{D}_{f}$ and preserve the knowledge in $\mathcal{D}_{r}$. 
Specifically, the optimization objective of \task\ is to minimize $\mathbb{E}_{(h, r, t) \sim \mathcal{D}_{r}, (h', r', t') \sim \mathcal{D}_{f}} [f(h', r', t') - f(h, r, t)]$.

\paragraph{Preference Optimization.}
Following~\cite{rafailov2024direct}, we formulate preference optimization as follows. 
Given a model $\mathcal{M}^{sft}$ trained by supervised fine-tuning, the answer pair $(y_{1}, y_{2}) \sim \mathcal{M}^{sft}(\cdot|x)$ can be generated by $\mathcal{M}^{sft}$ with each input $x$. 
The answer pair $(y_{1}, y_{2})$ will be labeled as $(y_{w}, y_{l})$ by human labels or reference policy model $\mathcal{M}^{ref}$, where $y_w$ and $y_l$ denote preferred and dis-preferred completion amongst $(y_{1}, y_{2})$, respectively. 
For the comparison dataset $\mathcal{D}_{c} = \{ (x^{(i)}, y_{w}^{(i)}, y_{l}^{(i)}) \}_{i=1}^{N} $, where $N$ denotes the number of $\mathcal{D}_{c}$, preference optimization aims to increase the output probability of $y_{w}^{i}$ and decrease the output probability of $y_{l}^{i}$. 
The objective of preference optimization is to minimize $\mathbb{E}_{(x, y_{w}, y_{l}) \sim \mathcal{D}_{c}}[r_{\phi}(x, y_{l}) - r_{\phi}(x, y_{w})]$, where $r_{\phi}$ denotes the training model initialized by $\mathcal{M}^{sft}$.

\subsection{Framework}
\begin{figure*}[t]
\centering
\includegraphics[width=1\textwidth]{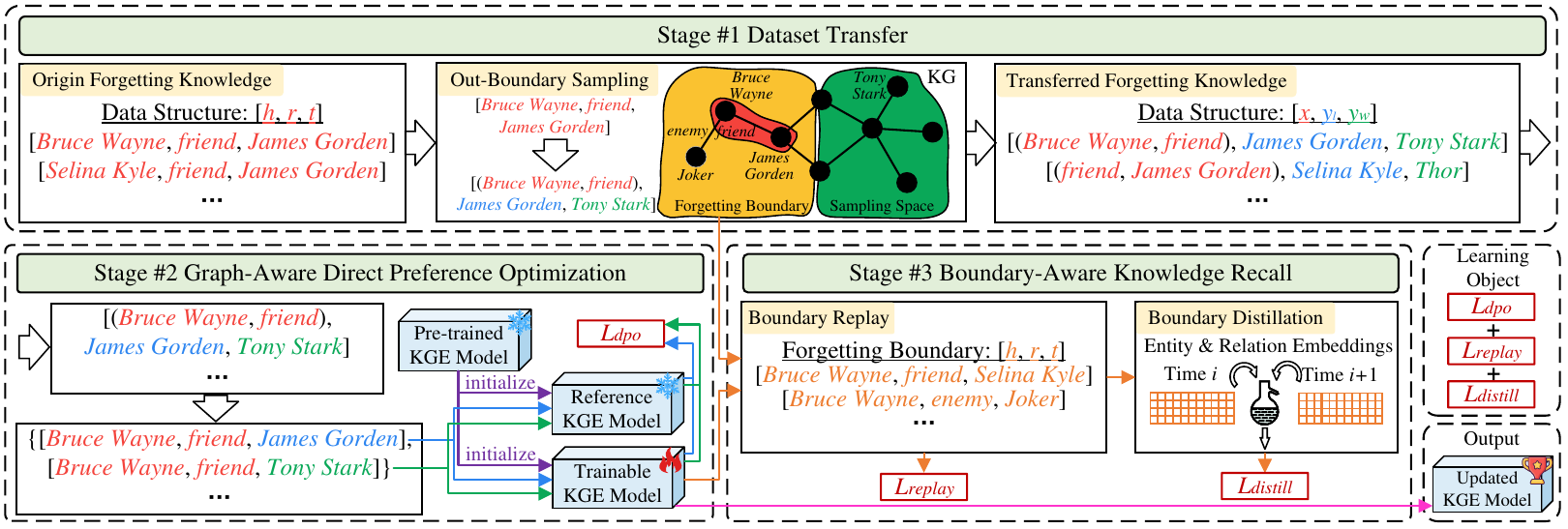} 
\vspace{-6mm}
\caption{An overview of \model\ framework.}
\vspace{-6mm}
\label{framework_picture}
\end{figure*}

The framework of \model\ is illustrated in Figure~\ref{framework_picture}. 
In Stage 1, we transfer the original forgetting knowledge using out-boundary sampling to datasets that are adaptive to preference optimization. 
In Stage 2, we apply the graph-aware direct preference optimization algorithm to achieve forgetting. 
In Stage 3, we design a boundary-aware knowledge recall method to retain the boundary knowledge.

\subsection{Graph-Aware Direct Preference Optimization}
To address the issue of forgotten knowledge being inferred from remaining knowledge, we reformulate unlearning as a preference optimization problem and employ a graph-aware direct preference optimization (DPO) algorithm with an out-boundary sampling strategy.

\paragraph{Dataset Transfer.}
To transfer unlearning in KGs to preference optimization, we extract the input $x$, the dis-preferred output $y_{l}$ and construct the preferred output $y_{w}$ required by the preference dataset from the forgetting dataset $\mathcal{D}_{f}$. 
Firstly, we extract each input $x$ and dis-preferred output $y_{l}$ from each sample of $\mathcal{D}_{f}$. 
Given the forgetting dataset $\mathcal{D}_{f} = \{ (h^{(i)}, r^{(i)}, t^{(i)}) \}_{i=1}^{N_{F}}$, where $N_{F}$ denotes the number of $\mathcal{D}_{f}$, we denote $D_{f}^{po}$ from $\mathcal{D}_{f}$ as follows:
\begin{equation}
    \mathcal{D}_{f}^{po} = \{ (h^{(i)}, r^{(i)}, t^{(i)}, s^{(i)}) \}_{i=1}^{N_{F}}, s^{(i)} \in \{1, 0\}
    \label{dfpo}
\end{equation}
where $(h^{(i)}, r^{(i)}, t^{(i)}) \in \mathcal{D}_{f}$, and $s^{(i)}$ is sampled from a Bernoulli distribution with $p(s) \sim Bernoulli(0.5)$. 
For each sample, when $s=1$, we take the head entity $h$ as the dis-preferred output $y_{l}$ inferred by the input $x=(r,t)$. 
Similarly, when $s=0$, we take the tail entity $t$ as the dis-preferred output $y_{l}$ inferred by the input $x=(h,r)$. 

Secondly, we construct the preferred output $y_{w}$ for each forgetting sample $(x, y_{l})$. 
We randomly sample an entity $e \in \mathcal{E}$ such that $e \neq y_{l}$ from the knowledge graph $\mathcal{G} = \{\mathcal{E}, \mathcal{R}, \mathcal{T}\}$ as a preferred output answer $y_{w}$ to the input $x$. 
The distribution of $y_{w}$ can be formally approximated as follows: 
\begin{equation}
    p(y_{w}) \cong p(y_{w}|x,y_{l}) = \frac{1-p(y_{l})}{|\mathcal{E}|-1}, \exists \hspace{0.1cm} y_{w} \in \mathcal{E} \wedge y_{w} \neq y_{l}
    \label{pyw}
\end{equation}
Thus, the final forgetting dataset $\mathcal{D}_{f}^{po}$ is defined as follows:
\begin{equation}
    \mathcal{D}_{f}^{po} = \{ (x^{(i)}, y_{l}^{(i)}, y_{w}^{(i)}) \}_{i=1}^{N_{F}}
    \label{final_dfpo}
\end{equation}
Meanwhile, we note that if $x=(h, r)$, then $(x, y_{l})$ and $(x, y_{w})$ correspond to the triples $(h, r, y_{l})$ and $(h, r, y_{w})$, respectively. 
Similarly, if $x=(r, t)$, then $(x, y_{l})$ and $(x, y_{w})$ correspond to the triples $(y_{l}, r, t)$ and $(y_{w}, r, t)$, respectively. 
Finally, we transfer the forgetting dataset $\mathcal{D}_{f}$ to $\mathcal{D}_{f}^{po}$. 
As shown in Figure~\ref{framework_picture} Stage 1, we transfer the forgetting knowledge from [\textit{Bruce Wayne, friend, James Gorden}] to [\textit{(Bruce Wayne, friend), James Gorden, Tony Stark}].

\paragraph{Task Transfer.}
In this section, we reformulate the unlearning task defined over the forgetting set $\mathcal{D}_f$ as a preference optimization problem over a constructed dataset $\mathcal{D}_{f}^{po}$. To support this formulation, we show that the two objectives are approximately equivalent to a linear transformation, thereby justifying the validity of the task transfer. 
We propose the following theorem:

\begin{theorem}[Equivalence of Optimization Objectives] Let $E_u$ be the expectation of the unlearning objective with respect to $\mathcal{D}_f$, and $E_p$ be the expectation of the preference objective with respect to $\mathcal{D}_f^{po}$, then we have:
\begin{equation}
\begin{aligned}
E_p = c_{1} \cdot E_{u} - c_{2}
\label{them1}
\end{aligned}
\end{equation}

where $c_1 > 1$ and $c_2 > 0$ are constants dependent on the candidate entity set size of $\mathcal{D}_f$. 
\label{Theme1}
\end{theorem}

See Appendix~\ref{appendix:proof-task-transfer} for a detailed proof.
Theorem~\ref{Theme1} demonstrates that minimizing the unlearning expectation $E_p$ is approximately equivalent to minimizing the preference expectation $E_u$.
Thus, preference optimization serves as a principled surrogate for the unlearning task.
Although $E_u$ and $E_p$ are not exactly equivalent in a strict distributional sense, their optimization objectives align closely, justifying that minimizing $E_p$ effectively minimizes $E_u$.

\paragraph{Direct Preference Optimization.}
After reformulating the unlearning task as a preference optimization problem, we apply preference optimization algorithms to address it. 
Inspired by DPO~\cite{rafailov2024direct}, we use the DPO algorithm to minimize the expectation of the preference objective $E_{p}$. 
Specifically, given the pre-trained KGE model $\mathcal{M}$, and the transferred forgetting dataset $\mathcal{D}_{f}^{po}$, we initialize the reference policy model $\mathcal{M}^{ref}$ using the parameters of $\mathcal{M}$. 
The loss for DPO is defined as follows, aiming to maximize $f_{\theta}(x, y_{w})$ and minimize $f_{\theta}(x, y_{l})$:
\begin{equation}
\begin{aligned}
\mathcal{L}_{dpo} = -\mathbb{E}_{(x,y_{w},y_{l}) \sim \mathcal{D}_{f}^{po}} [log\sigma(\beta log \frac{f_{\theta}(x, y_{w})}{f_{ref}(x, y_{w})} - \beta log \frac{f_{\theta}(x, y_{l})}{f_{ref}(x, y_{l})})]
\label{ldpo}
\end{aligned}
\end{equation}
where $f_{\theta}(\cdot)$ and $f_{ref}(\cdot) \in (0, 1)$ denote the score functions of $\mathcal{M}$ and $\mathcal{M}^{ref}$, respectively. 
The hyperparameter $\beta$ balances the preference and dis-preference data, and $\sigma(\cdot)$ represents the softmax function. 
We use TransE~\cite{bordes2013translating} as the base KGE model, where the score function is defined as $f(h, r, t) = Sigmoid( -|\mathbf{h} + \mathbf{r} - \mathbf{t}|_{L_{2}}) \in (0, 1)$, with $\mathbf{h}$, $\mathbf{r}$, and $\mathbf{t}$ denoting the embeddings of $h$, $r$, and $t$, respectively. 
Notice that the loss of DPO algorithm is equal to minimizing the preference optimization object $E_p$~\cite{rafailov2024direct}, we apply the reference optimization in practice. 
During training, the parameters of $\mathcal{M}^{ref}$ are frozen, and only the parameters of $\mathcal{M}$ are updated. 
This approach allows the model $\mathcal{M}$ to be optimized using the DPO algorithm. 
As shown in Figure~\ref{framework_picture} Stage 2, we use [\textit{Bruce Wayne, friend, Tony Stark}] as the preferred triple, and [\textit{Bruce Wayne, friend, James Gorden}] as the dis-preferred triple, which is fed into the reference model $\mathcal{M}^{ref}$ and trainable model $\mathcal{M}$.

\paragraph{Out-Boundary Sampling.}
To enhance the preference-based unlearning, we propose an out-boundary sampling strategy to improve the selection of $y_w$. 
Specifically, we avoid sampling $y_w$ from the close neighborhood of $y_l$, as neighboring entities tend to share similar embeddings~\cite{shangMultiGranularity}, leading to less discriminative preference signals.

We first we denote forgetting boundary $\mathcal{B}_{y_{l}}$ of $y_{l}$, and the boundary entities $\mathcal{E}_{y_{l}}$ as follows: 
\begin{equation} \mathcal{B}_{y_l} = \{ (h, r, t) \mid h = y_l \vee t = y_l \}, \quad \mathcal{E}_{y_l} = \{ e \mid (e, r, t) \in \mathcal{B}_{y_l} \vee (h, r, e) \in \mathcal{B}_{y_l} \}. 
\label{eq:boundary} 
\end{equation}
Since single-hop neighboring entities tend to have similar entity representations in KGEs~\cite{shangMultiGranularity}, 
if $y_{w}$ is drawn from $\mathcal{E}_{y_{l}}$, then $f_{\theta}(x, y_{w})$ in Equation~\ref{ldpo} tends to approximate $f_{\theta}(x, y_{l})$, making them hard to differentiate and optimize. 
Therefore, we modify the distribution of $y_{w}$ as: 
\begin{equation}
\begin{aligned}
    p_{o}(y_{w}) \cong p_{o}(y_{w}|x,y_{l}) &= \frac{1-p(y_{l})-p(\mathcal{E}_{y_{l}})}{|\mathcal{E}|-1-|\mathcal{E}_{y_{l}}|} \approx \frac{1-p(y_{l})-p(\mathcal{E}_{y_{l}})}{|\mathcal{E}|-1},
    \exists \hspace{0.1cm} y_{w} \in (\mathcal{E} \backslash \mathcal{E}_{yl}) , |\mathcal{E}| \gg |\mathcal{E}_{y_{l}}|
    \label{poyw}
\end{aligned}
\end{equation}
This way, preferred scores $f_{\theta}(x, y_{w})$ and dis-preferred scores $f_{\theta}(x, y_{l})$ are well separated by bout-boundary sampling, which promotes preference optimization. 
Finally, we demonstrate that out-boundary sampling does not alter the correspondence between the optimization objectives of the unlearning task and the preference optimization task. 
We propose the following theorem: 

\begin{theorem}[Preservation of Optimization Objective] Let $E_u$ be the expectation of the unlearning objective with respect to $\mathcal{D}_f$, and $E_p'$ be the modified preference objective under out-boundary sampling with respect to $\mathcal{D}_f^{po}$, then we have: 
\begin{equation}
E^{'}_p=c^{'}_1 \cdot E_u - c^{'}_2
\end{equation}

where $c_1' > 1$ and $c_2' > 0$ are constants dependent on the candidate entity set size of $\mathcal{D}_f$.

\label{Theme2}
\end{theorem}

See Appendix~\ref{appendix:obs-proof} for a detailed proof. 
From Theorem~\ref{Theme2}, we show that minimizing $E_p'$ remains approximately equivalent to minimizing the original unlearning objective $E_u$.

\subsection{Boundary-Aware Knowledge Recall}
In order to preserve remaining knowledge, \model\ introduces a boundary-aware knowledge recall mechanism to ensure minimal destruction of remaining knowledge. 
Specifically, since unlearning mainly affects the forgetting boundary~\cite{wang2023comprehensivesurveyforgettingdeep}, we replay and distill the forgetting boundary.

\paragraph{Boundary Replay.}
Specifically, we denote $\mathcal{D}_{replay}$ as the set of all triples from $\mathcal{B}_{y_l}$ for each dis-preferred entity $y_l$ in $\mathcal{D}_f^{po}$. 
The replay loss on $\mathcal{D}_{replay}$ is defined as follows: 
\begin{equation}
    \mathcal{L}_{replay} = \mathbb{E}_{(h, r, t) \sim \mathcal{D}_{replay}}   [max(0, f(h, r, t) - f(\overline{h}, r, \overline{t}) + \gamma)]
    \label{replayloss}
\end{equation}
where $(\overline{h}, r, \overline{t})$ is the negative triple of $(h, r, t)$ with random replacement of head and tail entities, and $\gamma$ is the margin.

\paragraph{Boundary Distillation.}
Inspired by recent work on knowledge distillation for KGEs~\cite{zhu2022dualde,liu2024towards}, we introduce a distillation strategy to reinforce knowledge retention around the forgetting boundary. Specifically, we define the set of distillation entities $\mathcal{E}_{distill}$ as all boundary neighbors of dis-preferred entities $y_l$ in $\mathcal{D}_f^{po}$, excluding $y_l$ itself. 
Following~\cite{liu2024towards}, we design the distillation loss for each entity $e \in \mathcal{E}_{distill}$ as follows: 
\begin{equation}
\begin{aligned}
\mathcal{L}_{e} = \left\{\begin{matrix} 
  \frac{1}{2}(\mathbf{e} - \mathbf{e}_{ref})^2, &|\mathbf{e} - \mathbf{e}_{ref}| \le 1 \\  
  |\mathbf{e} - \mathbf{e}_{ref}|-\frac{1}{2},&|\mathbf{e} - \mathbf{e}_{ref}| > 1
\end{matrix}\right. 
\end{aligned}
\end{equation}
where $\mathbf{e}$ and $\mathbf{e}_{ref}$ denote the embeddings for $e$ in the trained model $\mathcal{M}$ and the reference model $\mathcal{M}_{ref}$, respectively. 
Finally, we denote the distillation loss for $\mathcal{E}_{distill}$ as follows: 
\begin{equation}
    \mathcal{L}_{distill} = \mathbb{E}_{e \sim \mathcal{E}_{distill}} [\mathcal{L}_{e}]
\end{equation}

The final optimization object is to minimize $\mathcal{L}$ as follows:
\begin{equation}
    \mathcal{L} = \lambda_{1} \cdot \mathcal{L}_{dpo} + \lambda_{2} \cdot \mathcal{L}_{replay} + \lambda_{3} \cdot \mathcal{L}_{distill}
\end{equation}
where $\lambda_{1}$, $\lambda_{2}$, and $\lambda_{3}$ are the weights to balance losses.

\vspace{-2mm}
\section{Experiments}
\subsection{Experimental Setup}
\paragraph{Datasets.}
We construct eight datasets with 10\% and 20\% unlearning rate:  \textbf{FB-10\%}, \textbf{FB-20\%}, \textbf{WN-10\%}, \textbf{WN-20\%}, \textbf{CO-10\%}, \textbf{CO-20\%}, \textbf{YA-10\%}, and \textbf{YA-20\%}, which are based on four datasets with different scales: \textbf{FB15K-237}~\cite{dettmers2018convolutional}, \textbf{WN18RR}~\cite{toutanova2015representing}, \textbf{CoDEx-L}~\cite{safavi2020codex}, and \textbf{Yago3-10}~\cite{mahdisoltani2013yago3}. 
Each dataset has 4 unlearning time steps. 
At each time step $i$, we construct the forgetting dataset $\mathcal{D}_{f}^{i}$ and the remaining dataset $\mathcal{D}_{r}^{i}$. 
Details of data construction and specific statistics are shown in Appendix~\ref{appendix:dataset}. 

\vspace{-1mm}
\paragraph{Backbones and Baselines.}
We take \textbf{TransE}~\cite{bordes2013translating} as the base KGE model in experiments. 
In explorable experiments, we also explore other 4 different base KGE models (\textbf{TansD}~\cite{wang2014knowledge}, \textbf{ComplEx}~\cite{trouillon2016complex}, \textbf{SimplEx}~\cite{kazemi2018simple}, \textbf{RotatE}~\cite{sun2019rotate}) to verify the scalability of our method. 
We compare our methods with 10 baselines, including: 
(i) Two exact unlearning methods: \textbf{Re-Train} and \textbf{Fine-Tune}, which re-train and fine-tune pre-trained KGE models with the whole remaining data $\mathcal{D}_{r}^{i}$ at each time step $i$, respectively. 
(ii) Two traditional approximate unlearning methods: \textbf{RL}~\cite{golatkar2020eternal}, \textbf{Fisher}~\cite{golatkar2020eternal}, and six newest approximate unlearning methods: \textbf{BS}~\cite{chen2023boundary}, \textbf{NG}~\cite{yao2023large}, \textbf{SSD}~\cite{foster2024fast}, \textbf{ADV-IMP}~\cite{cha2024learning}, \textbf{Schema}~\cite{xiao2025knowledge}, and \textbf{MetaEU}~\cite{xu2024learn}. 
As one of the approximate unlearning method, our \textbf{GraphDPO} mainly compares with approximate methods. 
Details of the implementations list in Appendix~\ref{appendix:settings}.

\vspace{-1mm}
\paragraph{Metrics.}
We evaluate the performance of the model on the link prediction task and use the Mean Reciprocal Rank (MRR) as the evaluation metric. 
At each time step $i$, we test the model on both the accumulated forgetting dataset $\sum_{j=1}^{i}\mathcal{D}_{f}^{j}$ and the remaining dataset $\mathcal{D}_{r}^{i}$. 
To evaluate the unlearning performance and preserving performance together, we denote $MRR_{Avg}^{i} = \frac{MRR_{r}^{i} + (1 - MRR_{f}^{i})}{2}$ and $MRR_{F1}^{i} = \frac{2 \cdot MRR_{r}^{i} \cdot (1 - MRR_{f}^{i})}{MRR_{r}^{i} + (1 - MRR_{f}^{i})}$, where $MRR_{r}^{i}$ and $MRR_{f}^{i}$ denote the MRR metric on $\mathcal{D}_{r}^{i}$ and $\sum_{j=1}^{i}\mathcal{D}_{f}^{j}$, respectively. 
Higher $MRR_{Avg}^{i}$ and $MRR_{F1}^{i}$ indicate better performance.

\subsection{Results}

\begin{table*}[tb!] 
\centering 
\setlength{\tabcolsep}{0.9mm} 
\tiny
\vspace{-4mm}
\caption{Main experimental results on FB-10\%, FB-20\%, WN-10\%, and WN-20\%. The bold scores indicate the best results of approximate unlearning methods and underlined scores indicate the second best results in $M_{Avg}$ and $M_{F1}$. All results are the average of 5 runs and are presented in \% form.} 
\begin{tabular}{l|cccc|cccc|cccc|cccc} 
\toprule & \multicolumn{4}{c|}{Time 1} & \multicolumn{4}{c|}{Time 2} & \multicolumn{4}{c|}{Time 3} & \multicolumn{4}{c}{Time 4} \\ 
\bottomrule
$\mathcal{D}_{f}$: FB-10\% & $M_{f} \downarrow$ & $M_{r}$ & $M_{Avg}$ & $M_{F1}$ & $M_{f} \downarrow$ & $M_{r}$ & $M_{Avg}$ & $M_{F1}$ & $M_{f} \downarrow$ & $M_{r}$ & $M_{Avg}$ & $M_{F1}$ & $M_{f} \downarrow$ & $M_{r}$ & $M_{Avg}$ & $M_{F1}$ \\ 
\hline 
Re-Train & 0.203 & 0.257 & 0.527 & 0.388 & 0.204 & 0.267 & 0.532 & 0.400 & 0.201 & 0.275 & 0.537 & 0.409 & 0.201 & 0.287 & 0.543 & 0.422 \\ 
Fine-Tune & 0.239 & 0.277 & 0.519 & 0.406 & 0.242 & 0.277 & 0.518 & 0.406 & 0.236 & 0.282 & 0.523 & 0.411 & 0.224 & 0.285 & 0.531 & 0.417 \\ 
\hline 
RL & 0.072 & 0.082 & 0.505 & 0.151 & 0.077 & 0.082 & 0.503 & 0.151 & 0.092 & 0.095 & 0.502 & 0.172 & 0.077 & 0.081 & 0.502 & 0.149 \\ 
Fisher & 0.012 & 0.012 & 0.500 & 0.023 & 0.009 & 0.009 & 0.500 & 0.018 & 0.007 & 0.007 & 0.500 & 0.014 & 0.007 & 0.007 & 0.500 & 0.014 \\ 
BS & 0.066 & 0.075 & 0.505 & 0.139 & 0.067 & 0.072 & 0.503 & 0.134 & 0.073 & 0.077 & 0.502 & 0.142 & 0.076 & 0.080 & 0.502 & 0.147 \\ 
NG & 0.143 & 0.176 & \textbf{0.517} & 0.292 & 0.089 & 0.117 & \underline{0.512} & 0.207 & 0.053 & 0.074 & \underline{0.511} & 0.138 & 0.035 & 0.052 & \underline{0.509} & 0.099 \\ 
SSD & 0.140 & 0.160 & 0.509 & 0.269 & 0.140 & 0.156 & 0.508 & 0.265 & 0.135 & 0.149 & 0.507 & 0.254 & 0.132 & 0.146 & 0.507 & 0.250 \\ 
ADV-IMP & 0.256 & 0.257 & 0.500 & \textbf{0.382} & 0.227 & 0.226 & 0.500 & \textbf{0.350} & 0.199 & 0.199 & 0.499 & \underline{0.318} & 0.190 & 0.189 & 0.499 & \underline{0.307} \\ 
Schema & 0.174 & 0.176 & 0.501 & 0.290 & 0.177 & 0.174 & 0.499 & 0.287 & 0.184 & 0.187 & 0.502 & 0.304 & 0.174 & 0.179 & 0.503 & 0.294 \\
MetaEU & 0.167 & 0.163 & 0.498 & 0.273 & 0.164 & 0.165 & 0.501 & 0.276 & 0.175 & 0.177 & 0.501 & 0.291 & 0.167 & 0.166 & 0.500 & 0.277 \\
\textbf{GraphDPO} & 0.158 & 0.180 & \underline{0.511} & \underline{0.296} & 0.156 & 0.184 & \textbf{0.514} & \underline{0.302} & 0.154 & 0.196 & \textbf{0.521} & \textbf{0.319} & 0.152 & 0.207 & \textbf{0.527} & \textbf{0.333} \\ 
\bottomrule 
$\mathcal{D}_{f}$: FB-20\% & $M_{f} \downarrow$ & $M_{r}$ & $M_{Avg}$ & $M_{F1}$ & $M_{f} \downarrow$ & $M_{r}$ & $M_{Avg}$ & $M_{F1}$ & $M_{f} \downarrow$ & $M_{r}$ & $M_{Avg}$ & $M_{F1}$ & $M_{f} \downarrow$ & $M_{r}$ & $M_{Avg}$ & $M_{F1}$ \\ 
\hline 
Re-Train & 0.201 & 0.281 & 0.540 & 0.416 & 0.210 & 0.325 & 0.558 & 0.461 & 0.203 & 0.388 & 0.593 & 0.522 & 0.173 & 0.541 & 0.684 & 0.654 \\ 
Fine-Tune & 0.224 & 0.277 & 0.527 & 0.408 & 0.206 & 0.276 & 0.535 & 0.410 & 0.197 & 0.301 & 0.552 & 0.438 & 0.184 & 0.376 & 0.596 & 0.515 \\ 
\hline 
RL & 0.074 & 0.083 & 0.504 & 0.153 & 0.076 & 0.081 & 0.503 & 0.149 & 0.085 & 0.088 & 0.501 & 0.160 & 0.075 & 0.078 & 0.502 & 0.144 \\ 
Fisher & 0.012 & 0.012 & 0.500 & 0.024 & 0.009 & 0.010 & 0.500 & 0.012 & 0.009 & 0.009 & 0.500 & 0.017 & 0.008 & 0.008 & 0.500 & 0.016 \\ 
BS & 0.069 & 0.078 & 0.505 & 0.145 & 0.071 & 0.076 & 0.503 & 0.140 & 0.067 & 0.071 & 0.502 & 0.131 & 0.069 & 0.072 & 0.502 & 0.134 \\ 
NG & 0.086 & 0.106 & 0.510 & 0.189 & 0.036 & 0.053 & 0.509 & 0.100 & 0.019 & 0.029 & 0.505 & 0.057 & 0.014 & 0.020 & 0.503 & 0.039 \\ 
SSD & 0.157 & 0.178 & \underline{0.511} & 0.294 & 0.152 & 0.172 & \underline{0.510} & 0.285 & 0.143 & 0.164 & \underline{0.511} & 0.276 & 0.132 & 0.163 & \underline{0.516} & 0.274 \\ 
ADV-IMP & 0.287 & 0.195 & 0.454 & \underline{0.306} & 0.262 & 0.191 & 0.464 & \underline{0.303} & 0.224 & 0.182 & 0.479 & \underline{0.295} & 0.197 & 0.175 & 0.489 & 0.287 \\
Schema & 0.168 & 0.172 & 0.502 & 0.285 & 0.165 & 0.176 & 0.506 & 0.291 & 0.162 & 0.178 & 0.508 & 0.294 & 0.158 & 0.181 & 0.512 & 0.298 \\
MetaEU & 0.162 & 0.167 & 0.503 & 0.278 & 0.160 & 0.174 & 0.507 & 0.288 & 0.157 & 0.179 & \underline{0.511} & \underline{0.295} & 0.154 & 0.183 & 0.515 & \underline{0.301} \\
\textbf{GraphDPO} & 0.156 & 0.189 & \textbf{0.517} & \textbf{0.309} & 0.154 & 0.191 & \textbf{0.518} & \textbf{0.311} & 0.151 & 0.218 & \textbf{0.521} & \textbf{0.347} & 0.150 & 0.227 & \textbf{0.527} & \textbf{0.333} \\ 
\bottomrule 
$\mathcal{D}_{f}$: WN-10\% & $M_{f} \downarrow$ & $M_{r}$ & $M_{Avg}$ & $M_{F1}$ & $M_{f} \downarrow$ & $M_{r}$ & $M_{Avg}$ & $M_{F1}$ & $M_{f} \downarrow$ & $M_{r}$ & $M_{Avg}$ & $M_{F1}$ & $M_{f} \downarrow$ & $M_{r}$ & $M_{Avg}$ & $M_{F1}$ \\ 
\hline 
Re-Train & 0.182 & 0.596 & 0.707 & 0.689 & 0.165 & 0.609 & 0.722 & 0.704 & 0.146 & 0.630 & 0.742 & 0.725 & 0.167 & 0.648 & 0.761 & 0.744 \\ 
Fine-Tune & 0.222 & 0.554 & 0.666 & 0.647 & 0.252 & 0.501 & 0.625 & 0.600 & 0.162 & 0.422 & 0.630 & 0.562 & 0.171 & 0.429 & 0.629 & 0.566 \\ 
\hline 
RL & 0.100 & 0.118 & \underline{0.509} & 0.208 & 0.064 & 0.066 & 0.501 & 0.123 & 0.044 & 0.042 & 0.499 & 0.080 & 0.020 & 0.019 & 0.499 & 0.037 \\ 
Fisher & 0.022 & 0.014 & 0.496 & 0.028 & 0.022 & 0.010 & 0.493 & 0.017 & 0.021 & 0.006 & 0.492 & 0.013 & 0.020 & 0.004 & 0.492 & 0.008 \\ 
BS & 0.093 & 0.105 & 0.506 & 0.188 & 0.017 & 0.013 & 0.498 & 0.026 & 0.009 & 0.008 & 0.499 & 0.016 & 0.007 & 0.007 & 0.500 & 0.013 \\ 
NG & 0.510 & 0.361 & 0.426 & \underline{0.416} & 0.507 & 0.362 & 0.427 & \underline{0.417} & 0.499 & 0.361 & 0.431 & \underline{0.419} & 0.501 & 0.361 & 0.430 & \underline{0.419} \\ 
SSD & 0.156 & 0.190 & \textbf{0.513} & 0.309 & 0.063 & 0.150 & \textbf{0.544} & 0.259 & 0.061 & 0.116 & \underline{0.527} & 0.207 & 0.047 & 0.116 & \underline{0.534} & 0.206 \\ 
ADV-IMP & 0.176 & 0.173 & 0.498 & 0.286 & 0.059 & 0.058 & 0.500 & 0.109 & 0.052 & 0.051 & 0.499 & 0.097 & 0.046 & 0.047 & 0.502 & 0.088 \\ 
Schema & 0.286 & 0.224 & 0.469 & 0.341 & 0.292 & 0.217 & 0.465 & 0.332 & 0.284 & 0.222 & 0.469 & 0.339 & 0.312 & 0.242 & 0.465 & 0.358 \\
MetaEU & 0.292 & 0.211 & 0.460 & 0.325 & 0.302 & 0.204 & 0.451 & 0.316 & 0.297 & 0.211 & 0.457 & 0.325 & 0.324 & 0.254 & 0.465 & 0.369 \\
\textbf{GraphDPO} & 0.367 & 0.323 & 0.479 & \textbf{0.428} & 0.304 & 0.320 & \underline{0.508} & \textbf{0.438} & 0.283 & 0.340 & \textbf{0.528} & \textbf{0.461} & 0.376 & 0.272 & \textbf{0.552} & \textbf{0.496} \\ 
\bottomrule 
$\mathcal{D}_{f}$: WN-20\% & $M_{f} \downarrow$ & $M_{r}$ & $M_{Avg}$ & $M_{F1}$ & $M_{f} \downarrow$ & $M_{r}$ & $M_{Avg}$ & $M_{F1}$ & $M_{f} \downarrow$ & $M_{r}$ & $M_{Avg}$ & $M_{F1}$ & $M_{f} \downarrow$ & $M_{r}$ & $M_{Avg}$ & $M_{F1}$ \\ 
\hline 
Re-Train & 0.166 & 0.616 & 0.725 & 0.709 & 0.127 & 0.644 & 0.759 & 0.741 & 0.080 & 0.679 & 0.800 & 0.781 & 0.026 & 0.762 & 0.868 & 0.855 \\ 
Fine-Tune & 0.210 & 0.572 & 0.681 & 0.664 & 0.238 & 0.513 & 0.638 & 0.613 & 0.131 & 0.479 & 0.674 & 0.618 & 0.172 & 0.526 & 0.677 & 0.643 \\ 
\hline 
RL & 0.135 & 0.140 & 0.501 & 0.035 & 0.020 & 0.019 & 0.499 & 0.038 & 0.010 & 0.011 & 0.500 & 0.021 & 0.013 & 0.014 & 0.500 & 0.027 \\ 
Fisher & 0.024 & 0.022 & 0.499 & 0.042 & 0.006 & 0.005 & 0.499 & 0.009 & 0.004 & 0.005 & 0.500 & 0.010 & 0.004 & 0.002 & 0.499 & 0.004 \\ 
BS & 0.039 & 0.037 & 0.499 & 0.072 & 0.009 & 0.009 & 0.500 & 0.017 & 0.006 & 0.007 & 0.500 & 0.014 & 0.006 & 0.007 & 0.500 & 0.014 \\ 
NG & 0.513 & 0.407 & 0.447 & \underline{0.443} & 0.513 & 0.403 & 0.441 & \underline{0.445} & 0.515 & 0.403 & 0.444 & \underline{0.440} & 0.517 & 0.405 & 0.494 & \underline{0.441} \\ 
SSD & 0.058 & 0.160 & 0.519 & 0.085 & 0.057 & 0.124 & \underline{0.533} & 0.218 & 0.042 & 0.137 & 0.547 & 0.240 & 0.030 & 0.144 & 0.557 & 0.250 \\ 
ADV-IMP & 0.157 & 0.158 & 0.500 & 0.266 & 0.143 & 0.145 & 0.501 & 0.248 & 0.137 & 0.139 & 0.501 & 0.239 & 0.095 & 0.099 & 0.502 & 0.178 \\ 
Schema & 0.172 & 0.208 & 0.518 & 0.332 & 0.184 & 0.223 & 0.520 & 0.350 & 0.128 & 0.234 & \underline{0.553} & 0.369 & 0.101 & 0.294 & \underline{0.597} & 0.443 \\
MetaEU & 0.184 & 0.224 & \underline{0.520} & 0.352 & 0.193 & 0.234 & 0.521 & 0.363 & 0.134 & 0.219 & 0.543 & 0.350 & 0.087 & 0.259 & 0.586 & 0.404 \\
\textbf{GraphDPO} & 0.220 & 0.319 & \textbf{0.549} & \textbf{0.452} & 0.258 & 0.325 & \textbf{0.534} & \textbf{0.451} & 0.143 & 0.352 & \textbf{0.605} & \textbf{0.499} & 0.118 & 0.433 & \textbf{0.658} & \textbf{0.581} \\
\bottomrule
\end{tabular}
\vspace{-6mm}
\label{main_results}
\end{table*}

\paragraph{Main Results.}
Main results on FB-10\%, FB-20\%, WN-10\%, and WN-20\% are presented in Table~\ref{main_results}, and the results on CO-10\%, CO-20\%, YA-10\%, YA-20\% are shown in Appendix~\ref{appendix:main_results}. 
First, compared to the exact unlearning methods, Re-Train and Fine-Tune, \model\ achieves 93\%-99\%, 82\%-97\%, 68\%-88\%, and 69\%-97\% of their performance in $MRR_{Avg}$ across all datasets. 
Similarly, \model\ achieves 73\%-80\%, 63\%-81\%, 62\%-88\%, and 61\%-90\% of their performance in $MRR_{F1}$. 
It shows that \model\ with only partial data closely approximates the performance of full-data training.

Second, compared to approximate unlearning baselines, \model\ achieves the best $MRR_{Avg}$ and $MRR_{F1}$ on most datasets. 
On FB-20\% and WN-20\%, \model\ outperforms other methods by 0.1\%–10.1\% in $MRR_{Avg}$ and 0.1\%–14.0\% in $MRR_{F1}$, demonstrating its strength in forgetting large-scale knowledge while preserving core information. 
On FB-10\% and WN-10\%, where the forgetting rate is lower, \model\ remains competitive: slightly underperforming in Time 1 and 2 (-1.8\% $MRR_{Avg}$, -2.5\% $MRR_{F1}$ on average), but outperforming in Time 3 and 4 (+1.2\% $MRR_{Avg}$, +3.7\% $MRR_{F1}$). 
This indicates that while low forgetting rates limit immediate gains, \model\ maintains strong performance in continual unlearning.

Third, we observe an unbalace between forgetting and retention in some baselines: Fisher achieves strong forgetting (e.g., 1.2\% drop in $M_f$ on FB-20\%) but also incur similar retention loss (1.2\% drop in $M_r$). 
Conversely, methods like NG retain more knowledge but forget less, for example, $M_f$ remains high at 51.0\% and 51.3\% on WN-10\% and WN-20\%, respectively. 
In contrast, \model\ achieves a better balance, with average $MRR_{Avg}$ and $MRR_{F1}$ reaching 53.5\% and 41.1\%, respectively.

Fourth, we observe that on large-scale KGs (CO-10\%, CO-20\%, YA-10\%, YA-20\%), our GraphDPO consistently outperforms all baselines in both $MRR_{Avg}$ and $MRR_{F_1}$. 
It demonstrates not only the scalability of our approach to large KGs, but also its superior performance compared to existing methods. 
In particular, when compared to two KGE-specific unlearning methods, Schema and MetaEU, GraphDPO achieves gains of 1.0\%–6.2\% in $MRR_{Avg}$ and 2.3\%–8.2\% in $MRR_{F_1}$, respectively.

\begin{table*}[tb!]
\centering
\setlength{\tabcolsep}{1mm}
\tiny
\caption{Ablation experimental results on FB-20\% and WN-20\%. Replay, Dis, and O-S denote boundary replay, boundary distillation and out-boundary sampling, respectively. All results are the average of 5 runs.}
\begin{tabular}{l|cccc|cccc|cccc|cccc}
\toprule
 & \multicolumn{4}{c|}{Time 1} & \multicolumn{4}{c|}{Time 2} & \multicolumn{4}{c|}{Time 3} & \multicolumn{4}{c}{Time 4} \\
\bottomrule
$\mathcal{D}_{f}$: FB-20\% & $M_{f} \downarrow$ & $M_{r}$ & $M_{Avg}$ & $M_{F1}$ & $M_{f} \downarrow$ & $M_{r}$ & $M_{Avg}$ & $M_{F1}$ & $M_{f} \downarrow$ & $M_{r}$ & $M_{Avg}$ & $M_{F1}$ & $M_{f} \downarrow$ & $M_{r}$ & $M_{Avg}$ & $M_{F1}$ \\
\hline
GraphDPO  & 0.156 & 0.189 & \textbf{0.517} & \textbf{0.309} & 0.154 & 0.191 & \textbf{0.518} & \textbf{0.311} & 0.151 & 0.218 & \textbf{0.534} & \textbf{0.347} & 0.150 & 0.275 & \textbf{0.562} & \textbf{0.415} \\
\hline
w/o DPO   & 0.130 & 0.141 & \underline{0.505} & 0.242 & 0.153 & 0.176 & 0.511 & 0.230 & 0.164 & 0.204 & 0.520 & \underline{0.327} & 0.167 & 0.245 & 0.539 & \underline{0.378} \\
w/o Replay & 0.139 & 0.097 & 0.479 & 0.175 & 0.086 & 0.079 & 0.496 & 0.145 & 0.076 & 0.074 & 0.499 & 0.138 & 0.074 & 0.074 & 0.500 & 0.138 \\
w/o Dis   & 0.131 & 0.141 & \underline{0.505} & 0.242 & 0.118 & 0.154 & \underline{0.517} & 0.262 & 0.117 & 0.179 & \underline{0.529} & 0.297 & 0.119 & 0.221 & \underline{0.551} & 0.353 \\
w/o O-S   & 0.210 & 0.179 & 0.485 & \underline{0.292} & 0.168 & 0.160 & 0.496 & \underline{0.268} & 0.149 & 0.154 & 0.503 & 0.261 & 0.138 & 0.158 & 0.510 & 0.266 \\
\bottomrule
$\mathcal{D}_{f}$: WN-20\% & $M_{f} \downarrow$ & $M_{r}$ & $M_{Avg}$ & $M_{F1}$ & $M_{f} \downarrow$ & $M_{r}$ & $M_{Avg}$ & $M_{F1}$ & $M_{f} \downarrow$ & $M_{r}$ & $M_{Avg}$ & $M_{F1}$ & $M_{f} \downarrow$ & $M_{r}$ & $M_{Avg}$ & $M_{F1}$ \\
\hline
GraphDPO & 0.220 & 0.319 & \textbf{0.549} & \textbf{0.452} & 0.258 & 0.325 & \textbf{0.534} & \textbf{0.451} & 0.143 & 0.352 & \textbf{0.605} & \textbf{0.499} & 0.118 & 0.433 & \textbf{0.658} & \textbf{0.581} \\
\hline
w/o DPO   & 0.276 & 0.245 & \underline{0.485} & 0.366 & 0.282 & 0.302 & 0.510 & 0.425 & 0.284 & 0.384 & 0.549 & \underline{0.498} & 0.279 & \underline{0.485} & 0.603 & \underline{0.579} \\
w/o Replay & 0.443 & 0.273 & 0.415 & 0.366 & 0.303 & 0.156 & 0.427 & 0.255 & 0.079 & 0.038 & 0.479 & 0.072 & 0.067 & 0.032 & 0.482 & 0.061 \\
w/o Dis   & 0.360 & 0.278 & 0.459 & 0.388 & 0.286 & 0.308 & \underline{0.511} & 0.431 & 0.093 & 0.284 & \underline{0.596} & 0.433 & 0.121 & 0.394 & \underline{0.637} & 0.544 \\
w/o O-S   & 0.440 & 0.306 & 0.433 & \underline{0.395} & 0.359 & 0.338 & 0.490 & \underline{0.443} & 0.338 & 0.401 & 0.531 & \underline{0.498} & 0.207 & 0.396 & 0.595 & 0.528 \\
\bottomrule
\end{tabular}
\label{ablation_results}
\vspace{-2mm}
\end{table*}

\begin{figure}[t]
\centering
\includegraphics[width=1\textwidth]{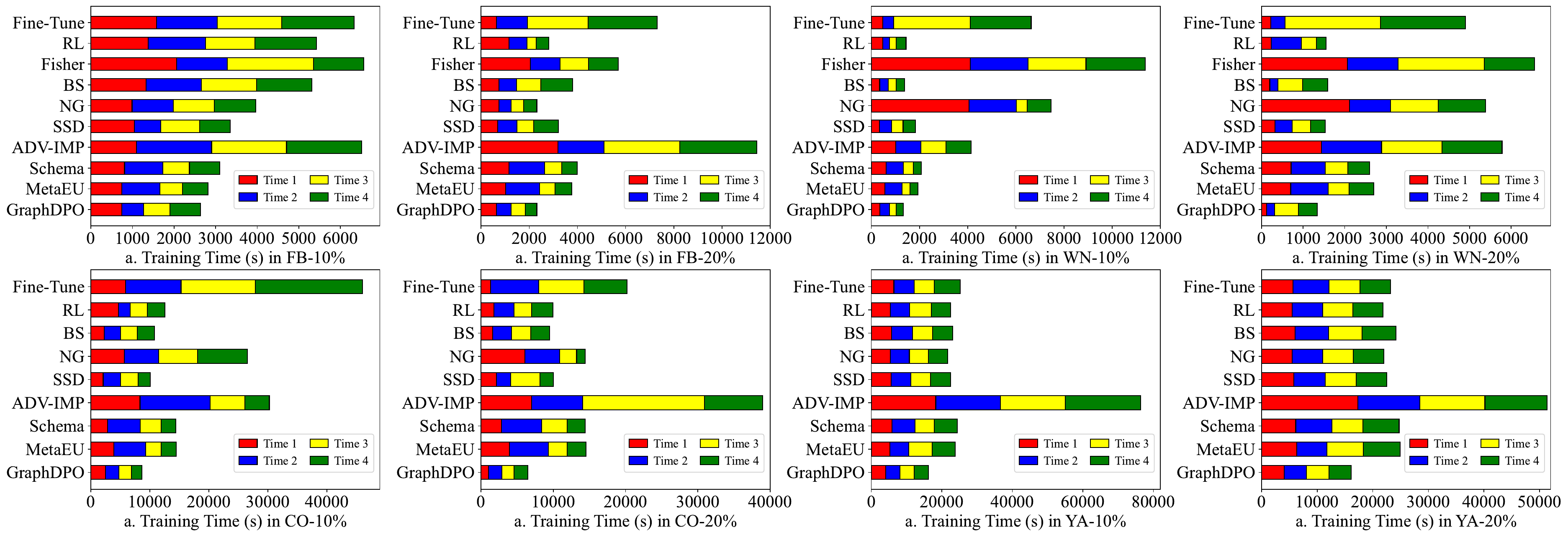}
\vspace{-6mm}
\caption{Time efficiency analysis. Fisher fails to run in the last four datasets with out-of-memory.}
\vspace{-6mm}
\label{time_efficiency}
\end {figure}

\begin{wrapfigure}{htb!}{0.5\textwidth}
\centering
\vspace{-4mm}
\includegraphics[width=0.5\textwidth]{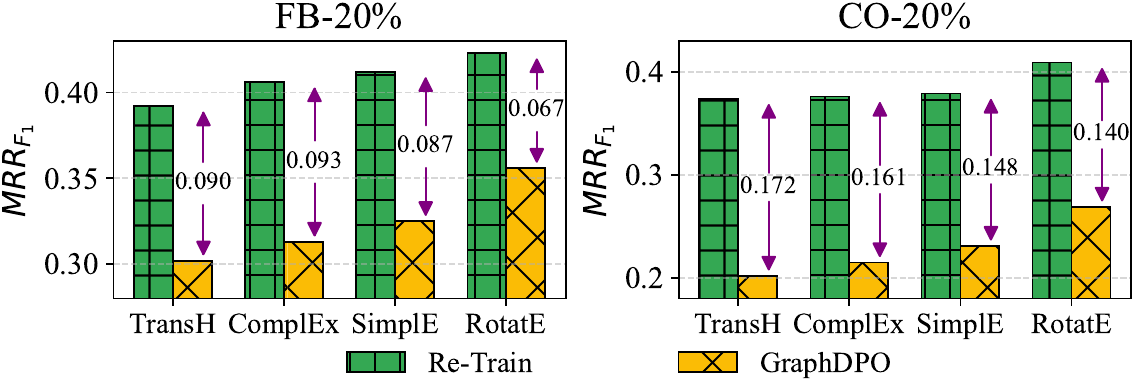}
\vspace{-6mm}
\caption{Scalability of different KGE models.}
\vspace{-6mm}
\label{kge_append}
\end{wrapfigure}

\paragraph{Ablation Results.}

To assess the effectiveness of each module in \model, we conduct ablation experiments on its full version and variants. 
Results on FB-20\% and WN-20\% are shown in Table~\ref{ablation_results}, with additional results in Appendix~\ref{appendix:ablation_results}. 
Removing any module leads to a performance drop of 0.1\%–17.6\% in $M_{Avg}$ and 0.2\%–52.0\% in $M_{F1}$, confirming their collective importance. 
In particular, removing the Replay module causes a 4.6\%–40.1\% drop in $M_r$, highlighting its key role in preserving retained knowledge. 
Removing the O-S module results in a 1.4\% and 12.6\% increase in $M_f$ on FB-20\% and WN-20\%, respectively, indicating its effectiveness in improving forgetting.

\begin{figure}[t]
\centering
\vspace{-4mm}
\includegraphics[width=1\textwidth]{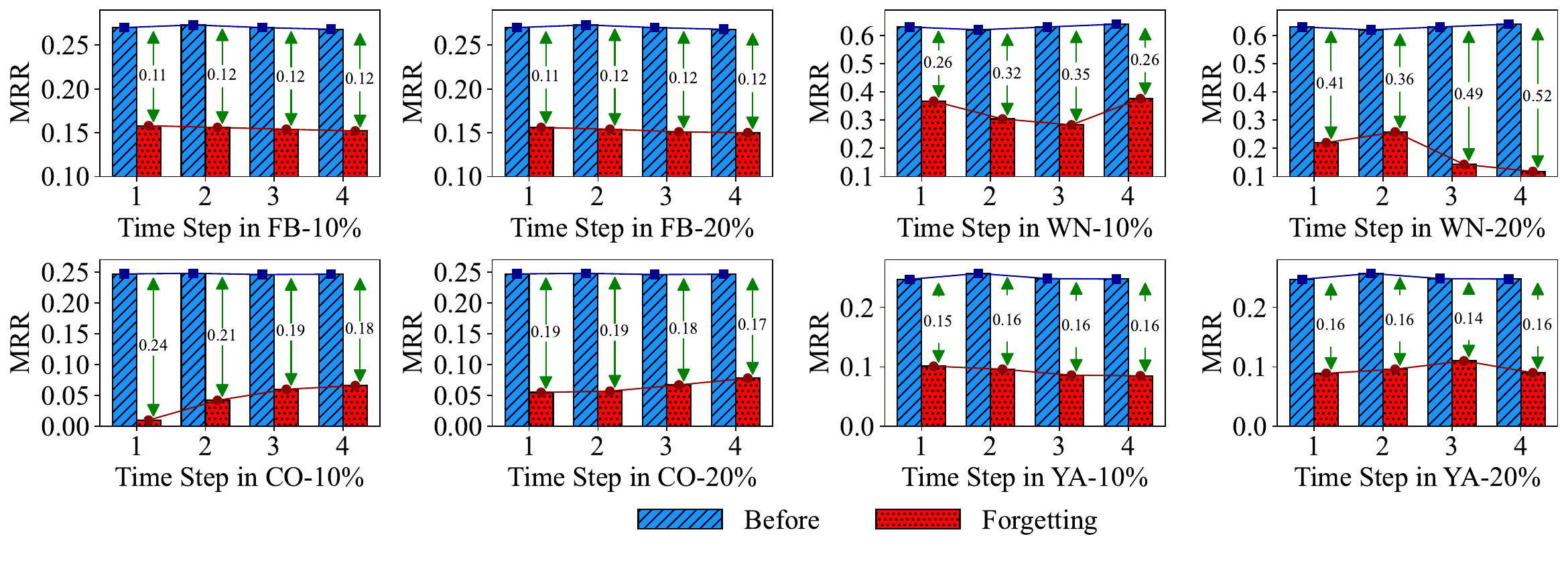}
\vspace{-6mm}
\caption{Forgetting analyse of \model. \textit{Before} defines the performance of original pre-trained model on $\mathcal{D}_{f}$, and \textit{Forgetting} defines the performance of GraphDPO on $\mathcal{D}_{f}$.}
\vspace{-2mm}
\label{forgetting_analyse}
\end {figure}

\begin{figure}[t]
\centering
\includegraphics[width=1\textwidth]{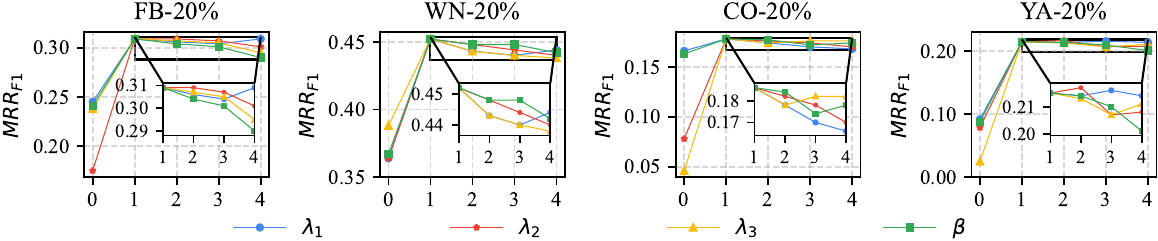}
\vspace{-4mm}
\caption{Hyperparametric analysis of $\lambda_1$, $\lambda_2$, $\lambda_3$, and $\beta$ with value from 0 to 4 in one time step.}
\vspace{-6mm}
\label{param_analyse}
\end {figure}

\paragraph{Time Efficiency.}
To evaluate the time efficiency, we compare the training time of all methods as shown in Figure~\ref{time_efficiency}. 
First, compared to the exact unlearning method Fine-Tune, \model\ saves 69\%-80\% of the training time on all datasets, highlighting its significant advantage over exact unlearning. 
Second, \model\ achieves the fastest training speed compared to approximate unlearning methods, reducing training time by 1\%-88\%.
It proves that \model\ is more efficient than other baselines.

\paragraph{Scalability of Different KGE Models.}
\label{backbones}
To assess the scalability of our method across different KGE models, we implement \model\ in TransH, ComplEx, SimplE, and RotatE on FB-20\% and CO-20\%, and compare it with full retraining in Figure~\ref{kge_append}. 
\model\ achieves 77\%–84\% of the full retraining performance on FB-20\% and 54\%–66\% on CO-20\%, demonstrating its effectiveness across various KGE backbones. 
Moreover, we notice that the performance gap between \model\ and retraining as the underlying KGE model improves (e.g., from 9.0\% to 6.07\% on FB-20\%, and 17.2\% to 14.0\% on CO-20\%), suggesting that \model\ benefits more from stronger base models.

\paragraph{Forgetting Analyse.}
To investigate the forgetting capability of \model, we compare the effectiveness of \model\ on forgetting datasets with original pre-trained models, as shown in Figure~\ref{forgetting_analyse}. 
First, compared to original models, performance decreases notably by 10\%-50\% on all datasets in $MRR_{f}$ after optimization by GraphDPO, demonstrating its significant forgetting performance. 
Second, the performance of \model\ on forgetting datasets decreases as time goes on, dropping from 0.16\% to 0.14\% and from 0.32\% to 0.13\%, reflecting its ability for continual forgetting.

\paragraph{Hyperparametric Analysis.}
To assess hyperparameter sensitivity, we select one time step from each of four datasets (FB-20\%, WN-20\%, CO-20\%, YA-20\%) and tune $\lambda_1$, $\lambda_2$, $\lambda_3$, and $\beta$. 
Results are shown in Figure~\ref{param_analyse}.  
First, we find that the performance drops notably when any hyperparameter is set to 0 and peaks at 1, indicating their necessity. 
Beyond 1, performance changes are minor (1\%–2\%), suggesting that our \model\ is robust and works well with default values set to 1.

\vspace{-2mm}
\begin{wrapfigure}{htb!}
{0.5\textwidth}
\centering
\small
\vspace{-4mm}
\includegraphics[width=0.43\textwidth]{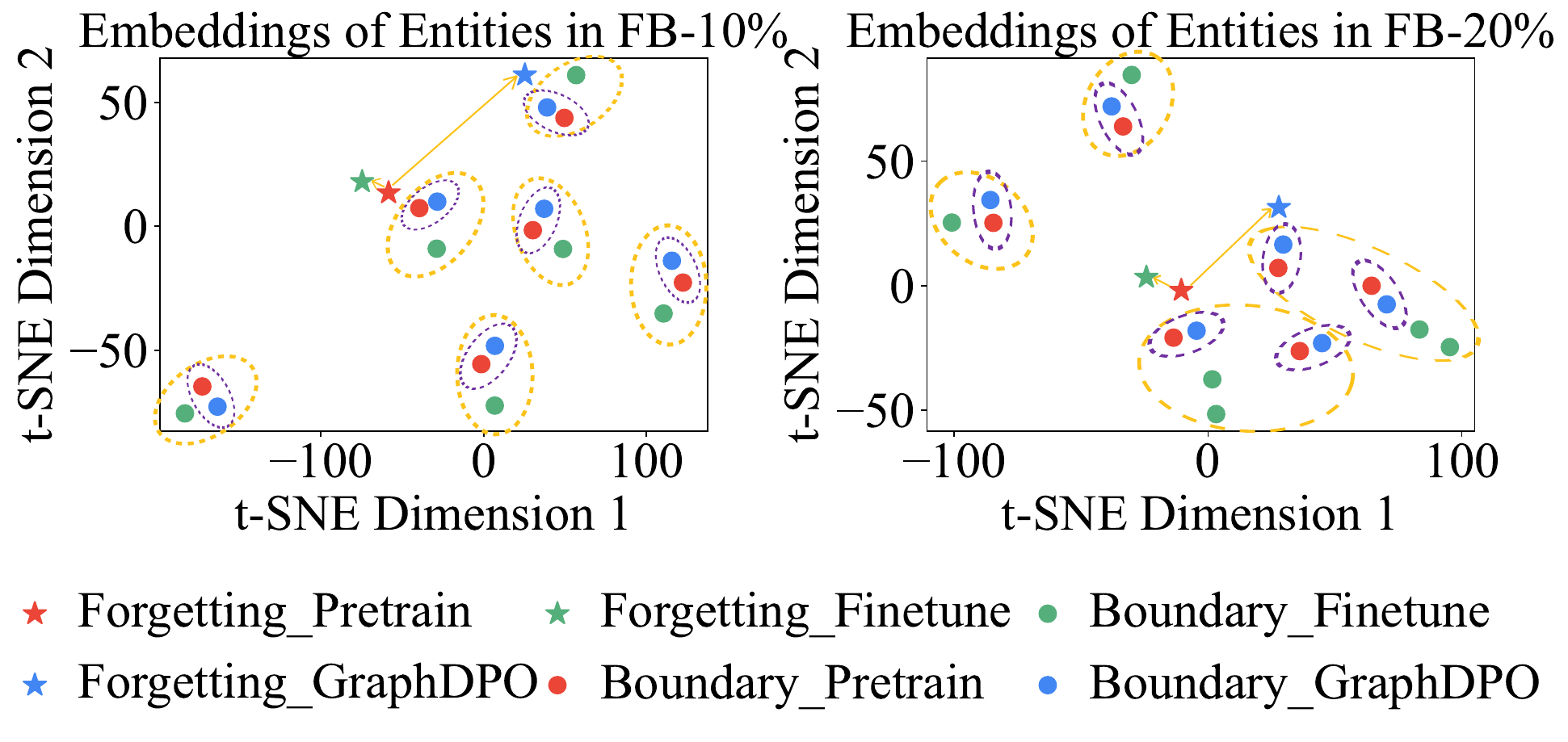}
\vspace{-4mm}
\caption{Visualization of forgetting and boundary entity embeddings.}
\vspace{-2mm}
\label{caseStudy}
\end{wrapfigure}

\paragraph{Visualization Analyse.}
To verify that our method can solve the issues of incomplete knowledge removal and the weakening of remaining knowledge, we use the t-SNE~\cite{van2008visualizing} to visualize the embeddings of both forgetting entities and those on the preserved boundary, as shown in Figure~\ref{caseStudy}. 
(1) For forgetting entities (marked by asterisks), \model\ makes a substantial shift in their embeddings compared to Finetune. 
This indicates that \model\ effectively addresses the incomplete knowledge removal problem. 
(2) For entities on the forgetting boundary (marked by circles), \model\ induces minimal changes in the embeddings relative to Finetune. 
It shows that \model\ can mitigate the destruction of the remaining knowledge.

\section{Conclusion}
In this paper, we dive into the task of unlearning for KGE. 
To address the challenge of forgetting knowledge in KGs due to their connectivity, we propose GraphDPO, which utilizes a graph-aware direct preference optimization algorithm with out-boundary sampling for effective forgetting learning. 
We also introduce boundary-aware knowledge recall to ensure that the remaining knowledge is preserved. 
In the future, we will explore more efficient unlearning methods on graphical models. 
More detailed limitations and future directions of unlearning in KGE are provided in Appendix~\ref{appendix:limitation}.

\bibliographystyle{unsrt}
\bibliography{ref}

\begin{thebibliography}{10}

\bibitem{wang2017knowledge}
Quan Wang, Zhendong Mao, Bin Wang, and Li~Guo.
\newblock Knowledge graph embedding: A survey of approaches and applications.
\newblock {\em IEEE Transactions on Knowledge and Data Engineering}, 29(12):2724--2743, 2017.

\bibitem{rossi2021knowledge}
Andrea Rossi, Denilson Barbosa, Donatella Firmani, Antonio Matinata, and Paolo Merialdo.
\newblock Knowledge graph embedding for link prediction: A comparative analysis.
\newblock {\em ACM Transactions on Knowledge Discovery from Data (TKDD)}, 15(2):1--49, 2021.

\bibitem{dong2014knowledge}
Xin Dong, Evgeniy Gabrilovich, Geremy Heitz, Wilko Horn, Ni~Lao, Kevin Murphy, Thomas Strohmann, Shaohua Sun, and Wei Zhang.
\newblock Knowledge vault: A web-scale approach to probabilistic knowledge fusion.
\newblock In {\em SIGKDD}, 2014.

\bibitem{bordes2014open}
Antoine Bordes, Jason Weston, and Nicolas Usunier.
\newblock Open question answering with weakly supervised embedding models.
\newblock In {\em ECML-PKDD}, 2014.

\bibitem{berant-liang-2014-semantic}
Jonathan Berant and Percy Liang.
\newblock Semantic parsing via paraphrasing.
\newblock In {\em ACL}, 2014.

\bibitem{barmettler2025conceptformer}
Joel Barmettler, Abraham Bernstein, and Luca Rossetto.
\newblock Conceptformer: Towards efficient use of knowledge-graph embeddings in large language models.
\newblock {\em arXiv preprint arXiv:2504.07624}, 2025.

\bibitem{said2023survey}
Anwar Said, Tyler Derr, Mudassir Shabbir, Waseem Abbas, and Xenofon Koutsoukos.
\newblock A survey of graph unlearning.
\newblock {\em arXiv preprint arXiv:2310.02164}, 2023.

\bibitem{10.1145/1242572.1242667}
Fabian~M. Suchanek, Gjergji Kasneci, and Gerhard Weikum.
\newblock Yago: A core of semantic knowledge.
\newblock In {\em WWW}, 2007.

\bibitem{HOFFART201328}
Johannes Hoffart, Fabian~M. Suchanek, Klaus Berberich, and Gerhard Weikum.
\newblock Yago2: A spatially and temporally enhanced knowledge base from wikipedia.
\newblock {\em Artificial Intelligence}, 2013.

\bibitem{cheng2024editing}
Siyuan Cheng, Ningyu Zhang, Bozhong Tian, Xi~Chen, Qingbin Liu, and Huajun Chen.
\newblock Editing language model-based knowledge graph embeddings.
\newblock In {\em AAAI}, 2024.

\bibitem{bourtoule2021machine}
Lucas Bourtoule, Varun Chandrasekaran, Christopher~A Choquette-Choo, Hengrui Jia, Adelin Travers, Baiwu Zhang, David Lie, and Nicolas Papernot.
\newblock Machine unlearning.
\newblock In {\em 2021 IEEE Symposium on Security and Privacy (SP)}, pages 141--159. IEEE, 2021.

\bibitem{eldan2023s}
Ronen Eldan and Mark Russinovich.
\newblock Who's harry potter? approximate unlearning in llms.
\newblock {\em arXiv preprint arXiv:2310.02238}, 2023.

\bibitem{wang2024machine}
Weiqi Wang, Zhiyi Tian, and Shui Yu.
\newblock Machine unlearning: A comprehensive survey.
\newblock {\em arXiv preprint arXiv:2405.07406}, 2024.

\bibitem{golatkar2020eternal}
Aditya Golatkar, Alessandro Achille, and Stefano Soatto.
\newblock Eternal sunshine of the spotless net: Selective forgetting in deep networks.
\newblock In {\em CVPR}, 2020.

\bibitem{thudi2022unrolling}
Anvith Thudi, Gabriel Deza, Varun Chandrasekaran, and Nicolas Papernot.
\newblock Unrolling sgd: Understanding factors influencing machine unlearning.
\newblock In {\em 2022 IEEE 7th European Symposium on Security and Privacy (EuroS\&P)}, pages 303--319. IEEE, 2022.

\bibitem{yan2022arcane}
Haonan Yan, Xiaoguang Li, Ziyao Guo, Hui Li, Fenghua Li, and Xiaodong Lin.
\newblock Arcane: An efficient architecture for exact machine unlearning.
\newblock In {\em IJCAI}, 2022.

\bibitem{wu2022puma}
Ga~Wu, Masoud Hashemi, and Christopher Srinivasa.
\newblock Puma: Performance unchanged model augmentation for training data removal.
\newblock In {\em AAAI}, 2022.

\bibitem{dettmers2018convolutional}
Tim Dettmers, Pasquale Minervini, Pontus Stenetorp, and Sebastian Riedel.
\newblock Convolutional 2d knowledge graph embeddings.
\newblock In {\em AAAI}, 2018.

\bibitem{toutanova2015representing}
Kristina Toutanova, Danqi Chen, Patrick Pantel, Hoifung Poon, Pallavi Choudhury, and Michael Gamon.
\newblock Representing text for joint embedding of text and knowledge bases.
\newblock In {\em EMNLP}, 2015.

\bibitem{safavi2020codex}
Tara Safavi and Danai Koutra.
\newblock Codex: A comprehensive knowledge graph completion benchmark.
\newblock In {\em EMNLP}, pages 8328--8350, 2020.

\bibitem{mahdisoltani2013yago3}
Farzaneh Mahdisoltani, Joanna Biega, and Fabian~M Suchanek.
\newblock Yago3: A knowledge base from multilingual wikipedias.
\newblock In {\em CIDR}, 2013.

\bibitem{nguyen2022survey}
Thanh~Tam Nguyen, Thanh~Trung Huynh, Phi~Le Nguyen, Alan Wee-Chung Liew, Hongzhi Yin, and Quoc Viet~Hung Nguyen.
\newblock A survey of machine unlearning.
\newblock {\em arXiv preprint arXiv:2209.02299}, 2022.

\bibitem{foster2024fast}
Jack Foster, Stefan Schoepf, and Alexandra Brintrup.
\newblock Fast machine unlearning without retraining through selective synaptic dampening.
\newblock In {\em AAAI}, 2024.

\bibitem{cha2024learning}
Sungmin Cha, Sungjun Cho, Dasol Hwang, Honglak Lee, Taesup Moon, and Moontae Lee.
\newblock Learning to unlearn: Instance-wise unlearning for pre-trained classifiers.
\newblock In {\em AAAI}, 2024.

\bibitem{xiao2025knowledge}
Yang Xiao, Ruimeng Ye, and Bo~Hui.
\newblock Knowledge graph unlearning with schema.
\newblock In {\em COLING}, pages 3541--3546, 2025.

\bibitem{xu2024learn}
Naixing Xu, Qian Li, Xu~Wang, Bingchen Liu, and Xin Li.
\newblock Learn to unlearn: Meta-learning-based knowledge graph embedding unlearning.
\newblock {\em arXiv preprint arXiv:2412.00881}, 2024.

\bibitem{bordes2013translating}
Antoine Bordes, Nicolas Usunier, Alberto Garcia-Duran, Jason Weston, and Oksana Yakhnenko.
\newblock Translating embeddings for modeling multi-relational data.
\newblock In {\em NIPS}, 2013.

\bibitem{trouillon2016complex}
Th{\'e}o Trouillon, Johannes Welbl, Sebastian Riedel, {\'E}ric Gaussier, and Guillaume Bouchard.
\newblock Complex embeddings for simple link prediction.
\newblock In {\em ICML}, 2016.

\bibitem{sun2019rotate}
Zhiqing Sun, Zhi-Hong Deng, Jian-Yun Nie, and Jian Tang.
\newblock Rotate: Knowledge graph embedding by relational rotation in complex space.
\newblock In {\em ICLR}, 2019.

\bibitem{bai2022training}
Yuntao Bai, Andy Jones, Kamal Ndousse, Amanda Askell, Anna Chen, Nova DasSarma, Dawn Drain, Stanislav Fort, Deep Ganguli, Tom Henighan, et~al.
\newblock Training a helpful and harmless assistant with reinforcement learning from human feedback.
\newblock {\em arXiv preprint arXiv:2204.05862}, 2022.

\bibitem{meng2024simpo}
Yu~Meng, Mengzhou Xia, and Danqi Chen.
\newblock Simpo: Simple preference optimization with a reference-free reward.
\newblock {\em arXiv preprint arXiv:2405.14734}, 2024.

\bibitem{schulman2017proximal}
John Schulman, Filip Wolski, Prafulla Dhariwal, Alec Radford, and Oleg Klimov.
\newblock Proximal policy optimization algorithms.
\newblock {\em arXiv preprint arXiv:1707.06347}, 2017.

\bibitem{rafailov2024direct}
Rafael Rafailov, Archit Sharma, Eric Mitchell, Christopher~D Manning, Stefano Ermon, and Chelsea Finn.
\newblock Direct preference optimization: Your language model is secretly a reward model.
\newblock In {\em NeurIPS}, 2024.

\bibitem{azar2024general}
Mohammad~Gheshlaghi Azar, Zhaohan~Daniel Guo, Bilal Piot, Remi Munos, Mark Rowland, Michal Valko, and Daniele Calandriello.
\newblock A general theoretical paradigm to understand learning from human preferences.
\newblock In {\em AISTATS}, 2024.

\bibitem{shangMultiGranularity}
Ziyu Shang, Peng Wang, Wenjun Ke, Jiajun Liu, Hailang Huang, Guozheng Li, Chenxiao Wu, Jianghan Liu, Xiye Chen, and Yining Li.
\newblock Learning multi-granularity and adaptive representation for knowledge graph reasoning.
\newblock In {\em IJCAI}, 2024.

\bibitem{wang2023comprehensivesurveyforgettingdeep}
Zhenyi Wang, Enneng Yang, Li~Shen, and Heng Huang.
\newblock A comprehensive survey of forgetting in deep learning beyond continual learning, 2023.

\bibitem{zhu2022dualde}
Yushan Zhu, Wen Zhang, Mingyang Chen, Hui Chen, Xu~Cheng, Wei Zhang, and Huajun Chen.
\newblock Dualde: Dually distilling knowledge graph embedding for faster and cheaper reasoning.
\newblock In {\em WSDM}, 2022.

\bibitem{liu2024towards}
Jiajun Liu, Wenjun Ke, Peng Wang, Ziyu Shang, Jinhua Gao, Guozheng Li, Ke~Ji, and Yanhe Liu.
\newblock Towards continual knowledge graph embedding via incremental distillation.
\newblock In {\em AAAI}, 2024.

\bibitem{wang2014knowledge}
Zhen Wang, Jianwen Zhang, Jianlin Feng, and Zheng Chen.
\newblock Knowledge graph embedding by translating on hyperplanes.
\newblock In {\em AAAI}, 2014.

\bibitem{kazemi2018simple}
Seyed~Mehran Kazemi and David Poole.
\newblock Simple embedding for link prediction in knowledge graphs.
\newblock {\em NeurIPS}, 2018.

\bibitem{chen2023boundary}
Min Chen, Weizhuo Gao, Gaoyang Liu, Kai Peng, and Chen Wang.
\newblock Boundary unlearning: Rapid forgetting of deep networks via shifting the decision boundary.
\newblock In {\em CVPR}, 2023.

\bibitem{yao2023large}
Yuanshun Yao, Xiaojun Xu, and Yang Liu.
\newblock Large language model unlearning.
\newblock {\em arXiv preprint arXiv:2310.10683}, 2023.

\bibitem{van2008visualizing}
Laurens Van~der Maaten and Geoffrey Hinton.
\newblock Visualizing data using t-sne.
\newblock {\em Journal of machine learning research}, 9(11), 2008.

\bibitem{lin2015learning}
Yankai Lin, Zhiyuan Liu, Maosong Sun, Yang Liu, and Xuan Zhu.
\newblock Learning entity and relation embeddings for knowledge graph completion.
\newblock In {\em AAAI}, 2015.

\bibitem{wang2021mulde}
Kai Wang, Yu~Liu, Qian Ma, and Quan~Z Sheng.
\newblock Mulde: Multi-teacher knowledge distillation for low-dimensional knowledge graph embeddings.
\newblock In {\em WWW}, pages 1716--1726, 2021.

\bibitem{liu2023iterde}
Jiajun Liu, Peng Wang, Ziyu Shang, and Chenxiao Wu.
\newblock Iterde: an iterative knowledge distillation framework for knowledge graph embeddings.
\newblock In {\em AAAI}, 2023.

\bibitem{chen2023entity}
Mingyang Chen, Wen Zhang, Zhen Yao, Yushan Zhu, Yang Gao, Jeff~Z Pan, and Huajun Chen.
\newblock Entity-agnostic representation learning for parameter-efficient knowledge graph embedding.
\newblock In {\em AAAI}, 2023.

\bibitem{daruna2021continual}
Angel Daruna, Mehul Gupta, Mohan Sridharan, and Sonia Chernova.
\newblock Continual learning of knowledge graph embeddings.
\newblock {\em IEEE Robotics and Automation Letters}, 6(2):1128--1135, 2021.

\bibitem{cui2023lifelong}
Yuanning Cui, Yuxin Wang, Zequn Sun, Wenqiang Liu, Yiqiao Jiang, Kexin Han, and Wei Hu.
\newblock Lifelong embedding learning and transfer for growing knowledge graphs.
\newblock In {\em AAAI}, 2023.

\bibitem{sinha2023distill}
Yash Sinha, Murari Mandal, and Mohan Kankanhalli.
\newblock Distill to delete: unlearning in graph networks with knowledge distillation.
\newblock {\em arXiv preprint arXiv:2309.16173}, 2023.

\bibitem{pan2023unlearning}
Chao Pan, Eli Chien, and Olgica Milenkovic.
\newblock Unlearning graph classifiers with limited data resources.
\newblock In {\em WWW}, pages 716--726, 2023.

\bibitem{zhang2024graph}
Jiahao Zhang.
\newblock Graph unlearning with efficient partial retraining.
\newblock In {\em Companion Proceedings of the ACM Web Conference 2024}, pages 1218--1221, 2024.

\bibitem{sinha2024distilldeleteunlearninggraph}
Yash Sinha, Murari Mandal, and Mohan Kankanhalli.
\newblock Distill to delete: Unlearning in graph networks with knowledge distillation, 2024.

\bibitem{lai2024step}
Xin Lai, Zhuotao Tian, Yukang Chen, Senqiao Yang, Xiangru Peng, and Jiaya Jia.
\newblock Step-dpo: Step-wise preference optimization for long-chain reasoning of llms.
\newblock {\em arXiv preprint arXiv:2406.18629}, 2024.

\bibitem{zeng2024token}
Yongcheng Zeng, Guoqing Liu, Weiyu Ma, Ning Yang, Haifeng Zhang, and Jun Wang.
\newblock Token-level direct preference optimization.
\newblock In {\em ICML}, pages 58348--58365, 2024.

\bibitem{gu2025mask}
Yuzhe Gu, Wenwei Zhang, Chengqi Lyu, Dahua Lin, and Kai Chen.
\newblock Mask-dpo: Generalizable fine-grained factuality alignment of llms.
\newblock {\em arXiv preprint arXiv:2503.02846}, 2025.

\bibitem{NEURIPS2019_9015}
Adam Paszke, Sam Gross, Francisco Massa, Adam Lerer, James Bradbury, and et~al.
\newblock Pytorch: An imperative style, high-performance deep learning library.
\newblock In {\em NeurIPS}, 2019.

\end{thebibliography}


\newpage
\appendix

\section{Detailed Related Work}
\label{appendix:related-work}

This section outlines the most relevant research of three key areas in detail: knowledge graph embedding, knowledge unlearning, and preference optimization. 

\paragraph{Knowledge Graph Embedding.} Knowledge graph embedding (KGE) aims to represent entities and relations in a continuous vector space while preserving the structural information of the graph. 
Early translational models such as TransE~\cite{bordes2013translating} interpret a relation as a translation vector between head and tail entities. 
Extensions like TransH~\cite{wang2014knowledge} and TransR~\cite{lin2015learning} introduce relation-specific hyperplanes or spaces to better model complex relations. 
Later, RotatE~\cite{sun2019rotate} represents relations as rotations in the complex vector space, capturing various relational patterns including symmetry and inversion. 
Recent studies have explored KGE methods under resource-constrained settings~\cite{wang2021mulde,zhu2022dualde,liu2023iterde,chen2023entity}. 
In parallel, continual and incremental knowledge graph embedding has gained attention, aiming to support dynamically evolving knowledge without retraining from scratch~\cite{daruna2021continual,cui2023lifelong,liu2024towards}. 
However, much less attention has been paid to the task of knowledge unlearning~\cite{xu2024learn,xiao2025knowledge}, which requires selectively forgetting specific facts or entities in the graph while preserving the rest of the knowledge. 
This work aims to bridge this gap by explicitly modeling unlearning in KGE. 

\paragraph{Knowledge Unlearning.} Knowledge unlearning aims to remove incorrect and outdated knowledge from machine learning models~\cite{wang2024machine}. 
Existing unlearning methods can be categorized into exact and approximate approaches. 
Exact unlearning methods~\cite{golatkar2020eternal,nguyen2022survey} require re-training the model using all reserved data, resulting in significant training costs. 
To mitigate this issue, approximate unlearning methods~\cite{thudi2022unrolling,wu2022puma} update models using only the forgetting datasets, with little or no use of the reserved datasets, thereby reducing training time~\cite{foster2024fast,cha2024learning}. 
Recently, several unlearning algorithms have been proposed for applications in graph learning~\cite{sinha2023distill,pan2023unlearning,zhang2024graph,sinha2024distilldeleteunlearninggraph}. 
Recent works utilize schema~\cite{xiao2025knowledge} or meta-learning~\cite{xu2024learn} to forget knowledge in KGE models, however, they struggle to effectively forget while preserving remaining knowledge~\cite{bordes2013translating,trouillon2016complex,sun2019rotate} due to the inherent connectivity of KGs.

\paragraph{Preference Optimization.} Preference optimization seeks to align LLMs with human preferences and values~\cite{bai2022training,lai2024step,zeng2024token,gu2025mask}, and can be categorized into online and offline algorithms~\cite{meng2024simpo}. 
Online algorithms incorporate reinforcement learning with supervised fine-tuning and policy optimization, which are inherently complex and challenging to optimize~\cite{schulman2017proximal}. 
To solve these issues, offline algorithms like DPO~\cite{rafailov2024direct} directly compare different decision sequences to optimize models, resulting in more efficient performance~\cite{azar2024general}. 
As alignment techniques leverage both positive and negative samples, offline algorithms are well-suited for selectively forgetting and remaining knowledge. 
In this paper, we demonstrate the effectiveness of applying DPO for unlearning in KGs.

\section{Proof of Theorem~1: Equivalence of Optimization Objectives}
\label{appendix:proof-task-transfer}

In this section, we prove the Equivalence of Minimizing $E_p$ and $E_u$. 
For the unlearning task based on the forgetting dataset $\mathcal{D}_f$, the optimization objective is to minimize $E_u$ as:
\begin{equation}
E_u = \mathbb{E}_{(h, r, t) \sim \mathcal{D}_f} [f(h, r, t)], \label{eq:appendix-eu} \end{equation}
where $h$, $r$, $t$ denote head entity, relation, and tail entity in a forgetting triple $(h,r,t)\in \mathcal{D}_f$, respectively. 
$f(\cdot)$ is the score function of KGE models.

For the preference optimization task based on $\mathcal{D}_f^{po}$, the optimization objective is to minimize $E_{p}$ as:
\begin{equation}
\begin{aligned}
E_{p} &= \mathbb{E}_{(x, y_{w}, y_{l}) \sim \mathcal{D}_{f}^{po}}[f(x, y_{l}) - f(x, y_{w})] \\
&= \mathbb{E}_{(x, y_{w}, y_{l}) \sim \mathcal{D}_{f}^{po}}[f(x, y_{l})] - \mathbb{E}_{(x, y_{w}, y_{l}) \sim \mathcal{D}_{f}^{po}}[f(x, y_{w})]
\label{Ep1}
\end{aligned}
\end{equation}
where $x$, $y_w$, and $y_l$ denote the query, prefered entity, and disprefered entity in $(x, y_w, y_l) \in \mathcal{D}_f^{po}$, respectively. 
For the first term in Equation~\ref{Ep1}, we have:
\begin{equation} 
\mathbb{E}_{(x, y_w, y_l) \sim \mathcal{D}_{f}^{po}} f(x, y_l) = \mathbb{E}_{(h, r, t) \sim \mathcal{D}_f}[f(h, r, t)] = E_u
\label{eq:appendix-eu-again} 
\end{equation}

For the second term in Equation~\ref{Ep1}, we analyze the expectation over the sampled $y_w$. 
Since $y_w$ is uniformly sampled from the candidate set $\mathcal{E} \setminus {t}$, we have: 

\begin{equation} \mathbb{E}_{(x, y_{w}, y_{l}) \sim \mathcal{D}_{f}^{po}}[f(x, y_{w})] = \frac{1}{|\mathcal{E}| - 1} [C - \mathbb{E}_{(x,y_w,y_l)\sim \mathcal{D}_{f}^{po}}f(x,y_l)], 
\label{eq:appendix-neg-estimate} 
\end{equation} 
where $C=\mathbb{E}_{(x,y_w,y_l)\in \mathcal{D}_f^{po}} \sum_{y \in \mathcal{E}} [f(x,y)]$. 
Then, we substitute Equation~\ref{eq:appendix-eu-again} and Equation~\ref{eq:appendix-neg-estimate} into Equation~\ref{Ep1}, we have: 

\begin{equation}
\begin{aligned}
E_{p} &= \mathbb{E}_{(x, y_{w}, y_{l}) \sim \mathcal{D}_{f}^{po}}f(x, y_{l}) - \frac{C - \mathbb{E}_{(x, y_{w}, y_{l}) \sim \mathcal{D}_{f}^{po}}f(x, y_{l})}{|\mathcal{E}|-1} \\
      &= \frac{|\mathcal{E}|}{|\mathcal{E}|-1} \cdot \mathbb{E}_{(x, y_{w}, y_{l}) \sim \mathcal{D}_{f}^{po}}f(x,y_{l}) - \frac{C}{|\mathcal{E}| - 1} \\
      &= c_{1} \cdot E_{u} - c_{2} 
\label{eq:appendix-ep1-final}
\end{aligned}
\end{equation}

where $c_1 = \frac{|\mathcal{E}|}{|\mathcal{E}| - 1} > 1$ is the positive constant, and $c_2 = \frac{C}{|\mathcal{E}| - 1} > 0$. 
Notice that $0 < f(x,y) < 1$, we have:
\begin{equation}
\begin{aligned}
0 &< c_2 = \frac{\mathbb{E}_{(x,y_w,y_l)\in \mathcal{D}_f^{po}} \sum_{y \in \mathcal{E}} [f(x,y)]}{|\mathcal{E}|-1}  <  \frac{\mathbb{E}_{(x,y_w,y_l)\in \mathcal{D}_f^{po}} |\mathcal{E}|}{|\mathcal{E}|-1}  =  \frac{|\mathcal{E}|}{\mathcal{|E|}-1} = c_1
\label{c2_boundary}
\end{aligned}
\end{equation}
Therefore, we have that $0 < c_2 < c_1$. 
From Equation~\ref{eq:appendix-ep1-final}, we observe that minimizing $E_{u}$ is approximately equivalent to minimizing $E_{p}$.

\section{Proof of Theorem~2: Preservation of Optimization Objective} 
\label{appendix:obs-proof}

In this section, we prove the equivalence of minimizing $E_p'$ and $E_u$. 
Combined with Equation~\ref{poyw}, the modified preference objective $E_p'$ can be calculated as: 
\begin{equation} 
\begin{aligned} 
E_p' &= \mathbb{E}_{(x, y_w, y_l) \sim \mathcal{D}_f^{po}} [f(x, y_l)] - \mathbb{E}_{(x, y_w, y_l) \sim \mathcal{D}_f^{po}} [f(x, y_w)] \\ 
     &= E_u - \frac{C - E_u - \sum_{e \in \mathcal{E}_{y_l}} \mathbb{E}_{(x,y_w,e)\sim\mathcal{D}_f^{po}}[f(x, e)]}{|\mathcal{E}| - 1} \\
     &= \frac{|\mathcal{E}|}{|\mathcal{E}| - 1} \cdot E_u - \frac{C - \sum_{e \in \mathcal{E}_{y_l}} \mathbb{E}_{(x,y_w,e)\sim\mathcal{D}_f^{po}}[f(x, e)]}{|\mathcal{E}| - 1} \\
     &= c_1' \cdot E_u - c_2'
\end{aligned} 
\end{equation}

where $C=\mathbb{E}_{(x,y_w,y_l)\in \mathcal{D}_f^{po}} \sum_{y \in \mathcal{E}} [f(x,y)]$, $c_1'=\frac{|\mathcal{E}|}{|\mathcal{E}|-1} > 1,
c_2'=\frac{C-\sum_{e \in \mathcal{E}_{y_l}}\mathbb{E}_{(x,y_w,e)\sim\mathcal{D}_f^{po}}[f(x,e)]}{|\mathcal{E}|-1}$. 
Since we have already shown in Equation~\ref{poyw} that $|\mathcal{E}| \gg |\mathcal{E}_{y_{l}}|$, and the values of $f(x, e)$ are bounded, the aggregated expectation over $\mathcal{E}_{y_l}$  becomes negligible compared to the total sum $C$.
Therefore, the correction term in $c_2'$ is very small, and we approximate:
\begin{equation} 
\begin{aligned} 
c_2' \approx \frac{C}{|\mathcal{E}| - 1}
\end{aligned} 
\end{equation}
Similary to Equation~\ref{c2_boundary}, we have:
\begin{equation}
\begin{aligned}
0 &< c_2' = \frac{\mathbb{E}_{(x,y_w,y_l)\in \mathcal{D}_f^{po}} \sum_{y \in \mathcal{E}} [f(x,y)]}{|\mathcal{E}|-1}  <  \frac{\mathbb{E}_{(x,y_w,y_l)\in \mathcal{D}_f^{po}} |\mathcal{E}|}{|\mathcal{E}|-1}  =  \frac{|\mathcal{E}|}{\mathcal{|E|}-1} = c_1'
\end{aligned}
\end{equation}
Therefore, we have that $0 < c_2' < c_1'$. 
Thus, $E_p' = c_1' \cdot E_u - c_2'$ still holds, and the equivalence is preserved.

\newpage

\section{Dataset Construction and Statistics}
\label{appendix:dataset}

\subsection{Dataset Construction}

We construct eight datasets based on four datasets \textbf{FB15K-237}~\cite{dettmers2018convolutional}, \textbf{WN18RR}~\cite{toutanova2015representing}, \textbf{CoDEx-L}~\cite{safavi2020codex}, and \textbf{Yago3-10}~\cite{mahdisoltani2013yago3}, which contains different scales of KGs, with 10\% and 20\% unlearning rate: \textbf{FB-10\%}, \textbf{FB-20\%}, \textbf{WN-10\%}, \textbf{WN-20\%}, \textbf{CO-10\%}, \textbf{CO-20\%}, \textbf{YA-10\%}, \textbf{YA-20\%}, which contain 10\% or 20\% triples that need to be forgotten. 
In order to further simulate the continuously evolving KGs in the real world, each dataset is set up with 4 time steps. 
At each time step $i$, we construct the forgetting dataset $\mathcal{D}_{f}^{i}$ and the remaining dataset $\mathcal{D}_{r}^{i}$. 
The $\mathcal{D}_{f}^{i}$ is drawn from the origin dataset $\mathcal{D}_{ori}$ by uniform sampling with the unlearning rate, and the $\mathcal{D}_{r}^{i} = \mathcal{D}_{ori} - \sum_{j=1}^{i} \mathcal{D}_{f}^{j}$. 

During dataset construction, at each time step, we sample both connected and unconnected triples with respect to the existing oblivion set. This ensures that each constructed oblivion set includes both triples that are semantically entangled with the retained knowledge, as well as those that are structurally independent.

The complete sampling procedure is illustrated in Algorithm~\ref{Algorithm1}, which incrementally builds the oblivion and remaining sets by iteratively identifying candidate triples, separating them by connectivity, and selecting a balanced subset at each step.

\begin{algorithm}[ht]
\label{Algorithm1}
\caption{Controlled Unlearning Dataset Construction}
\KwIn{Original triple set $\mathcal{D}_{\text{ori}}$, total unlearning rate $r$, number of time steps $T$}
\KwOut{Forgetting datasets $\{\mathcal{D}_f^1, \ldots, \mathcal{D}_f^T\}$ and remaining datasets $\{\mathcal{D}_r^1, \ldots, \mathcal{D}_r^T\}$}

Initialize historical forgetting set: $\mathcal{D}_{\text{hist}} \leftarrow \emptyset$\;

\For{$t = 1$ \KwTo $T$}{
    $\mathcal{C} \leftarrow \mathcal{D}_{\text{ori}} \setminus \mathcal{D}_{\text{hist}}$ \tcp*{\footnotesize Remaining candidates}
    $\mathcal{E}_{\text{hist}} \leftarrow$ all entities in $\mathcal{D}_{\text{hist}}$\;

    $\mathcal{C}_{\text{conn}} \leftarrow \{(h,r,t) \in \mathcal{C} \mid h \in \mathcal{E}_{\text{hist}} \vee t \in \mathcal{E}_{\text{hist}} \}$ \;
    $\mathcal{C}_{\text{unconn}} \leftarrow \mathcal{C} \setminus \mathcal{C}_{\text{conn}}$\;

    Sample $r \cdot |\mathcal{C}_{\text{conn}}|$ triples from $\mathcal{C}_{\text{conn}}$ as $\mathcal{D}_{f,\text{conn}}^t$\;
    Sample $r \cdot |\mathcal{C}_{\text{unconn}}|$ triples from $\mathcal{C}_{\text{unconn}}$ as $\mathcal{D}_{f,\text{unconn}}^t$\;

    $\mathcal{D}_f^t \leftarrow \mathcal{D}_{f,\text{conn}}^t \cup \mathcal{D}_{f,\text{unconn}}^t$\;
    $\mathcal{D}_r^t \leftarrow \mathcal{D}_{\text{ori}} \setminus \bigcup_{i=1}^{t} \mathcal{D}_f^i$\;

    $\mathcal{D}_{\text{hist}} \leftarrow \mathcal{D}_{\text{hist}} \cup \mathcal{D}_f^t$\;
}
\end{algorithm}

Compared to the datasets constructed in previous work~\cite{xiao2025knowledge,xu2024learn} for the unlearning of KGE, our datasets have the following advantages, which simulate different real-world situations:
(1) Covering various scales of KGs (from 92,583 triples to 1,089,000 triples). 
(2) Covering various connections between forgetting triples and remaining triples (forgetting triples of connecting and disconnecting to remaining triples both exist). 
(3) Covering various time steps, which simulates continual unlearning.

\subsection{Dataset Statistics}
The statistics of the origin datasets \textbf{FB15k-237}, \textbf{WN18RR}, \textbf{CoDEx-L}, and \textbf{YAGO3-10} are shown in Table~\ref{origin_dataset}.
The statistics of the constructed datasets \textbf{FB-10\%}, \textbf{FB-20\%}, \textbf{WN-10\%}, \textbf{WN-20\%}, \textbf{CO-10\%}, \textbf{CO-20\%}, \textbf{YA-10\%}, and \textbf{YA-20\%} are shown in Table~\ref{dataset}.

\begin{table}[ht]
\centering
\caption{The statistics of the origin datasets.}
\setlength{\tabcolsep}{8.5mm}
\begin{tabular}{cccc}
\toprule
Dataset & Entity Num & Relation Num & Triple Num \\
\midrule
FB15k-237     & 14,505     & 237   & 310,079 \\
WN18RR        & 40,559     & 11    & 92,583  \\
CoDEx-L       & 77,951     & 69    & 612,437 \\
YAGO3-10      & 123,143    & 37    & 1,089,000 \\
\bottomrule
\end{tabular}
\label{origin_dataset}
\end{table}

\begin{table*}[htb!]
\centering
\small
\caption{The statistics of the constructed datasets. $|\mathcal{E}|$, $|\mathcal{R}|$ and $|\mathcal{T}|$ denote the number of entities, relations and triples in the origin KG $\mathcal{D}_{ori}$. $|\mathcal{D}_{f}^{i}|$ and $|\mathcal{D}_{r}^{i}|$ denote the number of forgetting triples and remaining triples in $i$-th time step, respectively.}
\setlength{\tabcolsep}{0.7mm}
\begin{tabular}{lcccccccccccccc}
\toprule
\multirow{3}{*}{Dataset} & \multicolumn{3}{c}{$\mathcal{D}_{ori}$} & \multicolumn{2}{c}{Time 1} & \multicolumn{2}{c}{Time 2} & \multicolumn{2}{c}{Time 3} & \multicolumn{2}{c}{Time 4} \\
 & $|\mathcal{E}|$ & $|\mathcal{R}|$ & $|\mathcal{T}|$ & $|\mathcal{D}_{f}^{1}|$ & $|\mathcal{D}_{r}^{1}|$ & $|\mathcal{D}_{f}^{2}|$ & $|\mathcal{D}_{r}^{2}|$ & $|\mathcal{D}_{f}^{3}|$ & $|\mathcal{D}_{r}^{3}|$ & $|\mathcal{D}_{f}^{4}|$ & $|\mathcal{D}_{r}^{4}|$ \\
\hline
FB-10\% & 14,541 & 237 & 310,116 & 31,011 & 279,105 & 31,011 & 248,094 & 31,011 & 217,083 & 31,011 & 186,072 \\
FB-20\% & 14,541 & 237 & 310,116 & 62,023 & 248,093 & 62,023 & 186,070 & 62,023 & 124,047 & 62,023 & 62,024 \\
WN-10\% & 40,943 & 11 & 93,003 & 9,300 & 83,703 & 9,300 & 74,403 & 9,300 & 65,103 & 9,300 & 55,803 \\
WN-20\% & 40,943 & 11 & 93,003 & 18,600 & 74,403 & 18,600 & 55,803 & 18,600 & 37,203 & 18,600 & 18,603 \\
CO-10\% & 77,951 & 69 & 612,437 & 61,243 & 551,194 & 612,43 & 489,951 & 61,243 & 428,708 & 61,243 & 367,465 \\
CO-20\% & 77,951 & 69 & 612,437 & 122,487 & 489,950 & 122,487 & 367,463 & 122,487 & 244,976 & 122,487 & 122,489 \\
YA-10\% & 123,182 & 37 & 1,089,040 & 108,904 & 980,136 & 108,904 & 871,232 & 108,904 & 762,328 & 108,904 & 653,424 \\
YA-20\% & 123,182 & 37 & 1,089,040 & 217,808 & 871,232 & 217,808 & 653,424 & 127,808 & 435,616 & 217,808 & 217,808 \\
\bottomrule
\end{tabular}
\label{dataset}
\end{table*}

\section{Implementation Details}
\label{appendix:settings}

All experiments are conducted on a machine equipped with 4 NVIDIA RTX 3090Ti GPUs, utilizing the PyTorch framework~\cite{NEURIPS2019_9015}. 
As the base Knowledge Graph Embedding (KGE) model, we adopt TransE~\cite{bordes2013translating}, a well-established model for knowledge graph completion, to ensure a consistent baseline for comparison. 
Our method \model, and all baselines are implemented using the same KGE model architecture, which is pre-trained on the same original knowledge graph (KG) to ensure fairness in the experimental setup. 
The embedding size is 200, and the margin $\gamma$ for KGE pretraining is 8. 
For all baselines, we use the Adam as the optimizer. 
We tune the learning rate from the set [1e-3, 5e-3, 1e-2] and explore different batch sizes from [512, 1024] to identify the best configuration for convergence. For \model, we set the regularization parameters $\beta = \lambda_{1} = \lambda_{2} = \lambda_{3} = 1$ based on preliminary experiments, ensuring a balanced trade-off between different loss components. 
Additionally, we set the hyperparameter $\gamma = 8$ to control the margin in the distance-based loss function. 
To avoid replaying an excessive number of triples, we limit the replay dataset size $|\mathcal{D}_{replay}|$ to 10\% of the full dataset. Finally, all experimental results are reported as the average performance over five independent runs to ensure statistical reliability.

\section{Limitations}
\label{appendix:limitation}
While our method \model\ demonstrates strong performance across diverse KGE models and unlearning settings, one limitation is its reliance on pre-defined unlearning rates (e.g., 10\% or 20\%) when constructing the forgetting datasets. 
In practical scenarios, the scope and quantity of knowledge to be unlearned may be uncertain, context-dependent, or evolve over time, making fixed-rate strategies less flexible. 
Future work could explore adaptive mechanisms that estimate forgetting targets dynamically, such as leveraging confidence scores, usage frequency, or external feedback to determine both when and how much to forget in a more data-driven manner.

\newpage
\section{Main Results on CO-10\%, CO-20\%, YA-10\%, and YA-20\%}
\label{appendix:main_results}

\begin{table*}[htb!] 
\centering 
\setlength{\tabcolsep}{0.9mm} 
\tiny
\caption{Main experimental results on CO-10\%, CO-20\%, YA-10\%, and YA-20\%. The bold scores indicate the best results of approximate unlearning methods and underlined scores indicate the second best results in $M_{Avg}$ and $M_{F1}$. All results are the average of 5 runs and are presented in \% form. OOM denotes out of memory with 4 NVIDIA RTX 3090Ti GPUs.} 
\begin{tabular}{l|cccc|cccc|cccc|cccc} 
\toprule & \multicolumn{4}{c|}{Time 1} & \multicolumn{4}{c|}{Time 2} & \multicolumn{4}{c|}{Time 3} & \multicolumn{4}{c}{Time 4} \\ 
\bottomrule
$\mathcal{D}_{f}$: CO-10\% & $M_{f} \downarrow$ & $M_{r}$ & $M_{Avg}$ & $M_{F1}$ & $M_{f} \downarrow$ & $M_{r}$ & $M_{Avg}$ & $M_{F1}$ & $M_{f} \downarrow$ & $M_{r}$ & $M_{Avg}$ & $M_{F1}$ & $M_{f} \downarrow$ & $M_{r}$ & $M_{Avg}$ & $M_{F1}$ \\ 
\hline 
Re-Train & 0.134 & 0.229 & 0.548 & 0.362 & 0.128 & 0.238 & 0.555 & 0.373 & 0.122 & 0.249 & 0.563 & 0.388 & 0.142 & 0.299 & 0.579 & 0.444 \\
Fine-Tune & 0.141 & 0.216 & 0.537 & 0.345 & 0.138 & 0.230 & 0.546 & 0.363 & 0.136 & 0.249 & 0.556 & 0.387 & 0.132 & 0.278 & 0.573 & 0.421 \\
\hline 
RL & 0.040 & 0.051 & 0.506 & 0.097 & 0.049 & 0.058 & 0.504 & 0.109 & 0.053 & 0.059 & 0.503 & 0.111 & 0.051 & 0.056 & 0.502 & 0.106 \\
Fisher & OOM & OOM & OOM & OOM & OOM & OOM & OOM & OOM & OOM & OOM & OOM & OOM & OOM & OOM & OOM & OOM \\
BS & 0.025 & 0.031 & 0.503 & 0.061 & 0.035 & 0.040 & 0.503 & 0.078 & 0.044 & 0.049 & 0.502 & 0.092 & 0.045 & 0.048 & 0.502 & 0.092 \\
NG & 0.031 & 0.084 & \underline{0.526} & 0.154 & 0.031 & 0.075 & \underline{0.522} & 0.139 & 0.028 & 0.057 & \underline{0.515} & 0.107 & 0.013 & 0.051 & \underline{0.519} & 0.096 \\
SSD & 0.114 & 0.103 & 0.495 & \underline{0.184} & 0.106 & 0.101 & 0.498 & \underline{0.182} & 0.099 & 0.078 & 0.489 & 0.143 & 0.079 & 0.094 & 0.507 & \underline{0.170} \\
ADV-IMP & 0.091 & 0.093 & 0.501 & 0.169 & 0.077 & 0.079 & 0.501 & 0.146 & 0.080 & 0.070 & 0.495 & 0.129 & 0.049 & 0.079 & 0.515 & 0.146 \\
Schema & 0.102 & 0.097 & 0.497 & 0.175 & 0.091 & 0.086 & 0.498 & 0.158 & 0.087 & 0.071 & 0.492 & 0.132 & 0.071 & 0.061 & 0.495 & 0.115 \\
MetaEU & 0.111 & 0.087 & 0.488 & 0.158 & 0.107 & 0.074 & 0.484 & 0.137 & 0.097 & 0.086 & 0.494 & \underline{0.157} & 0.084 & 0.077 & 0.497 & 0.142 \\
\textbf{GraphDPO} & 0.010 & 0.110 & \textbf{0.550} & \textbf{0.197} & 0.042 & 0.101 & \textbf{0.529} & \textbf{0.183} & 0.060 & 0.092 & \textbf{0.516} & \textbf{0.167} & 0.066 & 0.108 & \textbf{0.521} & \textbf{0.194} \\
\bottomrule 
$\mathcal{D}_{f}$: CO-20\% & $M_{f} \downarrow$ & $M_{r}$ & $M_{Avg}$ & $M_{F1}$ & $M_{f} \downarrow$ & $M_{r}$ & $M_{Avg}$ & $M_{F1}$ & $M_{f} \downarrow$ & $M_{r}$ & $M_{Avg}$ & $M_{F1}$ & $M_{f} \downarrow$ & $M_{r}$ & $M_{Avg}$ & $M_{F1}$ \\ 
\hline 
Re-Train & 0.129 & 0.235 & 0.553 & 0.370 & 0.138 & 0.289 & 0.576 & 0.433 & 0.132 & 0.391 & 0.629 & 0.539 & 0.089 & 0.552 & 0.732 & 0.688 \\
Fine-Tune & 0.132 & 0.214 & 0.541 & 0.344 & 0.131 & 0.274 & 0.572 & 0.417 & 0.124 & 0.348 & 0.612 & 0.498 & 0.118 & 0.448 & 0.665 & 0.594 \\
\hline 
RL & 0.047 & 0.057 & 0.505 & 0.108 & 0.048 & 0.054 & 0.503 & 0.103 & 0.050 & 0.055 & 0.502 & 0.103 & 0.050 & 0.053 & 0.502 & 0.100 \\
Fisher & OOM & OOM & OOM & OOM & OOM & OOM & OOM & OOM & OOM & OOM & OOM & OOM & OOM & OOM & OOM & OOM \\
BS & 0.040 & 0.049 & 0.504 & 0.094 & 0.048 & 0.053 & 0.503 & 0.101 & 0.049 & 0.053 & 0.502 & 0.101 & 0.049 & 0.051 & 0.501 & 0.098 \\
NG & 0.042 & 0.081 & \underline{0.519} & 0.149 & 0.032 & 0.071 & \underline{0.520} & 0.133 & 0.027 & 0.068 & \underline{0.521} & 0.128 & 0.043 & 0.067 & \underline{0.512} & 0.126 \\
SSD & 0.115 & 0.095 & 0.490 & \underline{0.171} & 0.095 & 0.083 & 0.494 & 0.152 & 0.085 & 0.081 & 0.498 & 0.150 & 0.073 & 0.077 & 0.502 & 0.142 \\
ADV-IMP & 0.111 & 0.091 & 0.490 & 0.166 & 0.101 & 0.090 & 0.495 & \underline{0.164} & 0.091 & 0.081 & 0.495 & 0.149 & 0.084 & 0.075 & 0.495 & 0.138 \\
Schema & 0.083 & 0.077 & 0.497 & 0.141 & 0.085 & 0.080 & 0.497 & 0.146 & 0.094 & 0.086 & 0.496 & \underline{0.158} & 0.108 & 0.116 & 0.504 & \underline{0.206} \\
MetaEU & 0.100 & 0.076 & 0.488 & 0.141 & 0.113 & 0.084 & 0.485 & 0.154 & 0.110 & 0.080 & 0.485 & 0.147 & 0.093 & 0.084 & 0.495 & 0.154 \\
\textbf{GraphDPO} & 0.055 & 0.098 & \textbf{0.521} & \textbf{0.178} & 0.057 & 0.099 & \textbf{0.521} & \textbf{0.179} & 0.067 & 0.111 & \textbf{0.522} & \textbf{0.199} & 0.078 & 0.171 & \textbf{0.546} & \textbf{0.288} \\
\bottomrule 
$\mathcal{D}_{f}$: YA-10\% & $M_{f} \downarrow$ & $M_{r}$ & $M_{Avg}$ & $M_{F1}$ & $M_{f} \downarrow$ & $M_{r}$ & $M_{Avg}$ & $M_{F1}$ & $M_{f} \downarrow$ & $M_{r}$ & $M_{Avg}$ & $M_{F1}$ & $M_{f} \downarrow$ & $M_{r}$ & $M_{Avg}$ & $M_{F1}$ \\ 
\hline 
Re-Train & 0.091 & 0.158 & 0.533 & 0.269 & 0.091 & 0.168 & 0.539 & 0.284 & 0.090 & 0.181 & 0.545 & 0.302 & 0.091 & 0.203 & 0.556 & 0.332 \\
Fine-Tune & 0.079 & 0.123 & 0.522 & 0.217 & 0.078 & 0.133 & 0.527 & 0.233 & 0.080 & 0.144 & 0.532 & 0.249 & 0.080 & 0.156 & 0.538 & 0.267 \\
\hline 
RL & 0.021 & 0.028 & 0.504 & 0.055 & 0.025 & 0.030 & 0.502 & 0.057 & 0.025 & 0.029 & 0.502 & 0.056 & 0.026 & 0.029 & 0.501 & 0.057 \\
Fisher & OOM & OOM & OOM & OOM & OOM & OOM & OOM & OOM & OOM & OOM & OOM & OOM & OOM & OOM & OOM & OOM \\
BS & 0.011 & 0.015 & 0.502 & 0.029 & 0.019 & 0.022 & 0.502 & 0.043 & 0.024 & 0.027 & 0.502 & 0.053 & 0.024 & 0.026 & 0.501 & 0.051 \\
NG & 0.124 & 0.125 & 0.501 & \underline{0.218} & 0.119 & 0.119 & 0.500 & \underline{0.210} & 0.073 & 0.113 & \underline{0.520} & \underline{0.201} & 0.114 & 0.113 & 0.499 & \underline{0.200} \\
SSD & 0.060 & 0.087 & \underline{0.513} & 0.159 & 0.062 & 0.081 & \underline{0.510} & 0.149 & 0.040 & 0.079 & \underline{0.520} & 0.145 & 0.057 & 0.073 & \underline{0.508} & 0.136 \\
ADV-IMP & 0.114 & 0.115 & 0.500 & 0.203 & 0.113 & 0.113 & 0.500 & 0.200 & 0.054 & 0.053 & 0.500 & 0.101 & 0.045 & 0.045 & 0.500 & 0.086 \\
Schema & 0.121 & 0.106 & 0.493 & 0.190 & 0.121 & 0.119 & 0.499 & 0.209 & 0.115 & 0.110 & 0.498 & 0.196 & 0.103 & 0.109 & 0.503 & 0.195 \\
MetaEU & 0.130 & 0.088 & 0.479 & 0.160 & 0.110 & 0.074 & 0.482 & 0.137 & 0.110 & 0.068 & 0.479 & 0.126 & 0.104 & 0.057 & 0.477 & 0.108 \\
\textbf{GraphDPO} & 0.101 & 0.130 & \textbf{0.514} & \textbf{0.227} & 0.096 & 0.132 & \textbf{0.518} & \textbf{0.231} & 0.086 & 0.128 & \textbf{0.521} & \textbf{0.225} & 0.085 & 0.124 & \textbf{0.520} & \textbf{0.218} \\
\bottomrule 
$\mathcal{D}_{f}$: YA-20\% & $M_{f} \downarrow$ & $M_{r}$ & $M_{Avg}$ & $M_{F1}$ & $M_{f} \downarrow$ & $M_{r}$ & $M_{Avg}$ & $M_{F1}$ & $M_{f} \downarrow$ & $M_{r}$ & $M_{Avg}$ & $M_{F1}$ & $M_{f} \downarrow$ & $M_{r}$ & $M_{Avg}$ & $M_{F1}$ \\ 
\hline 
Re-Train & 0.092 & 0.167 & 0.537 & 0.282 & 0.091 & 0.202 & 0.555 & 0.330 & 0.094 & 0.268 & 0.587 & 0.414 & 0.097 & 0.501 & 0.702 & 0.644 \\ 
Fine-Tune & 0.079 & 0.131 & 0.526 & 0.229 & 0.077 & 0.151 & 0.537 & 0.259 & 0.080 & 0.183 & 0.551 & 0.305 & 0.086 & 0.234 & 0.574 & 0.372 \\ 
\hline 
RL & 0.023 & 0.030 & 0.503 & 0.058 & 0.025 & 0.029 & 0.502 & 0.056 & 0.024 & 0.026 & 0.501 & 0.051 & 0.023 & 0.025 & 0.501 & 0.049 \\ 
Fisher & OOM & OOM & OOM & OOM & OOM & OOM & OOM & OOM & OOM & OOM & OOM & OOM & OOM & OOM & OOM & OOM \\
BS & 0.020 & 0.025 & 0.503 & 0.049 & 0.024 & 0.028 & 0.502 & 0.054 & 0.025 & 0.027 & 0.501 & 0.052 & 0.025 & 0.026 & 0.501 & 0.051 \\ 
NG & 0.139 & 0.119 & 0.490 & \underline{0.209} & 0.127 & 0.136 & 0.504 & \underline{0.235} & 0.124 & 0.130 & 0.503 & \underline{0.226} & 0.115 & 0.121 & 0.503 & \underline{0.212} \\ 
SSD & 0.057 & 0.080 & \underline{0.511} & 0.147 & 0.058 & 0.077 & 0.509 & 0.142 & 0.057 & 0.077 & 0.510 & 0.142 & 0.054 & 0.081 & \underline{0.514} & 0.150 \\ 
ADV-IMP & 0.076 & 0.076 & 0.500 & 0.141 & 0.070 & 0.070 & 0.500 & 0.131 & 0.063 & 0.064 & 0.500 & 0.119 & 0.047 & 0.048 & 0.500 & 0.091 \\ 
Schema & 0.091 & 0.102 & 0.506 & 0.184 & 0.087 & 0.109 & \underline{0.511} & 0.195 & 0.092 & 0.121 & \underline{0.515} & 0.214 & 0.110 & 0.097 & 0.494 & 0.175 \\
MetaEU & 0.097 & 0.097 & 0.500 & 0.175 & 0.103 & 0.092 & 0.495 & 0.167 & 0.107 & 0.102 & 0.498 & 0.183 & 0.118 & 0.084 & 0.483 & 0.153 \\
\textbf{GraphDPO} & 0.089 & 0.122 & \textbf{0.517} & \textbf{0.215} & 0.096 & 0.138 & \textbf{0.521} & \textbf{0.239} & 0.110 & 0.166 & \textbf{0.528} & \textbf{0.279} & 0.090 & 0.126 & \textbf{0.518} & \textbf{0.221} \\ 
\bottomrule
\end{tabular} 
\label{app_main_results}
\vspace{-4mm}
\end{table*}

\section{Ablation Results on CO-20\% and YA-20\%}
\label{appendix:ablation_results}

\begin{table*}[htb!] 
\centering 
\setlength{\tabcolsep}{1mm} 
\tiny
\caption{Ablation experimental results on CO-20\% and YA-20\%. Replay, Dis, and O-S denote boundary replay, boundary distillation and out-boundary sampling, respectively. All results are the average of 5 runs and are presented in \% form.} 
\begin{tabular}{l|cccc|cccc|cccc|cccc} 
\toprule & \multicolumn{4}{c|}{Time 1} & \multicolumn{4}{c|}{Time 2} & \multicolumn{4}{c|}{Time 3} & \multicolumn{4}{c}{Time 4} \\ 
\bottomrule
$\mathcal{D}_{f}$: CO-20\% & $M_{f} \downarrow$ & $M_{r}$ & $M_{Avg}$ & $M_{F1}$ & $M_{f} \downarrow$ & $M_{r}$ & $M_{Avg}$ & $M_{F1}$ & $M_{f} \downarrow$ & $M_{r}$ & $M_{Avg}$ & $M_{F1}$ & $M_{f} \downarrow$ & $M_{r}$ & $M_{Avg}$ & $M_{F1}$ \\ 
\hline 
GraphDPO & 0.055 & 0.098 & \textbf{0.521} & \textbf{0.178} & 0.057 & 0.099 & \textbf{0.521} & \textbf{0.179} & 0.067 & 0.111 & \textbf{0.522} & \textbf{0.199} & 0.078 & 0.171 & \textbf{0.546} & \textbf{0.288} \\ 
\hline 
w/o DPO & 0.074 & 0.091 & \underline{0.508} & \underline{0.165} & 0.081 & 0.091 & 0.505 & \underline{0.165} & 0.085 & 0.104 & 0.510 & 0.187 & 0.083 & 0.151 & 0.534 & 0.259 \\ 
w/o Replay & 0.100 & 0.041 & 0.470 & 0.078 & 0.042 & 0.055 & 0.507 & 0.105 & 0.061 & 0.099 & \underline{0.519} & 0.180 & 0.073 & 0.158 & \underline{0.542} & 0.270 \\ 
w/o Dis & 0.052 & 0.024 & 0.486 & 0.046 & 0.036 & 0.019 & 0.492 & 0.038 & 0.028 & 0.018 & 0.495 & 0.036 & 0.026 & 0.019 & 0.497 & 0.038 \\ 
w/o O-S & 0.060 & 0.061 & 0.500 & 0.114 & 0.068 & 0.088 & \underline{0.510} & 0.161 & 0.076 & 0.106 & 0.515 & \underline{0.190} & 0.086 & 0.167 & 0.541 & \underline{0.282} \\ 
\bottomrule $\mathcal{D}_{f}$: YA-20\% & $M_{f} \downarrow$ & $M_{r}$ & $M_{Avg}$ & $M_{F1}$ & $M_{f} \downarrow$ & $M_{r}$ & $M_{Avg}$ & $M_{F1}$ & $M_{f} \downarrow$ & $M_{r}$ & $M_{Avg}$ & $M_{F1}$ & $M_{f} \downarrow$ & $M_{r}$ & $M_{Avg}$ & $M_{F1}$ \\ 
\hline 
GraphDPO & 0.089 & 0.122 & \textbf{0.517} & \textbf{0.215} & 0.096 & 0.138 & \textbf{0.521} & \textbf{0.239} & 0.110 & 0.166 & \textbf{0.528} & \textbf{0.279} & 0.090 & 0.126 & \textbf{0.518} & \textbf{0.221}  \\ 
\hline 
w/o DPO & 0.138 & 0.048 & 0.455 & \underline{0.091} & 0.147 & 0.067 & 0.460 & \underline{0.124} & 0.136 & 0.082 & 0.473 & \underline{0.150} & 0.138 & 0.118 & 0.490 & \underline{0.208} \\ 
w/o Replay & 0.132 & 0.041 & 0.455 & 0.079 & 0.057 & 0.050 & 0.497 & 0.094 & 0.049 & 0.063 & 0.507 & 0.119 & 0.048 & 0.090 & 0.521 & 0.164 \\ 
w/o Dis & 0.025 & 0.013 & 0.494 & 0.025 & 0.020 & 0.010 & 0.495 & 0.020 & 0.021 & 0.014 & 0.496 & 0.027 & 0.022 & 0.017 & 0.497 & 0.032 \\ 
w/o O-S & 0.026 & 0.029 & \underline{0.501} & 0.057 & 0.037 & 0.048 & \underline{0.505} & 0.091 & 0.044 & 0.065 & \underline{0.511} & 0.122 & 0.047 & 0.082 & \underline{0.517} & 0.151 \\ 
\bottomrule 
\end{tabular} 
\label{app_ablation_results} 
\end{table*}

\end{document}